\documentclass[fleqn,10pt]{wlscirep}
\usepackage[utf8]{inputenc}
\usepackage[T1]{fontenc}
\usepackage{graphicx}
\usepackage{amsmath}
\usepackage{amssymb}

\usepackage{subfiles}
\usepackage{hyperref}
\usepackage{xurl}
\usepackage[most]{tcolorbox}
\usepackage{tabularx}
\usepackage[table]{xcolor}
\usepackage{multirow}
\usepackage{makecell}
\usepackage{booktabs}

\usepackage{algorithm}
\usepackage{algpseudocode}

\usepackage[most]{tcolorbox}
\tcbuselibrary{breakable,listings}
\usepackage{xcolor}
\usepackage{enumitem}

\usepackage{listings}
\usepackage{xcolor}
\usepackage{longtable}
\usepackage{placeins}

\lstdefinestyle{jsonstyle}{
    basicstyle=\ttfamily\footnotesize,
    backgroundcolor=\color{white},
    frame=single,
    rulecolor=\color{gray!45},
    frameround=tttt,
    xleftmargin=0.5em,
    xrightmargin=0.5em,
    aboveskip=0.4em,
    belowskip=0.4em,
    columns=fullflexible,
    keepspaces=true,
    showstringspaces=false,
    breaklines=true
}

\title{Human Behavior-Informed Crash Scenario Generation with Real-World Crash Priors for Autonomous Vehicle Safety Evaluation}

\author[1]{Mingxing Peng}
\author[1]{Xusen Guo}
\author[2]{Long Chen}
\author[3]{Xintao Yan}
\author[4]{Siyu Teng}
\author[1,*]{Jun Ma}

\affil[1]{The Hong Kong University of Science and Technology (Guangzhou), China}
\affil[2]{Institute of Automation Chinese Academy of Sciences, Beijing, China}
\affil[3]{The University of Hong Kong, Hong Kong, China}
\affil[4]{Shenzhen University, Shenzhen, China}

\affil[*]{corresponding author: jun.ma@ust.hk}

\begin{abstract}
Reliable safety evaluation of autonomous vehicles (AVs) is essential to improving road safety, yet it depends critically on realistic simulation of rare crashes. Existing crash scenario generation methods can increase collision occurrence, but often fail to realistically reproduce how crashes evolve before impact or the distribution of crash types observed in the real world. Here, we present CrashSim, a human behavior-informed crash scenario generation framework that uses real-world crash priors to guide generative multi-agent traffic simulation for more reliable AV safety evaluation. These priors capture how real-world crashes evolve before impact and how different crash types are distributed, allowing limited crash data to guide realistic and scalable scenario generation across naturalistic driving contexts. We evaluate CrashSim against competing methods, showing that it more closely reproduces real-world pre-impact behavior, collision dynamics, collision geometry and crash-type distributions. We further use CrashSim to construct nuCrash dataset, containing over 4,000 crash and near-crash scenarios. Closed-loop evaluation of five AV planners shows that nuCrash more effectively exposes differences in planner safety capabilities than nuScenes. An LLM-assisted evaluation agent further analyzes planner failures to provide capability-level diagnoses and targeted improvement guidance. Together, CrashSim enables realistic and scalable crash generation for more informative AV safety evaluation.
\\


\end{abstract}

\begin{document}

\flushbottom
\maketitle

\rfoot{\small\sffamily\bfseries\thepage/\pageref{LastMainPage}}

\section*{Introduction}
\label{sec1}
\begin{figure}[!t]
\centering
\includegraphics[trim={70 475 125 110}, clip, width=\linewidth]{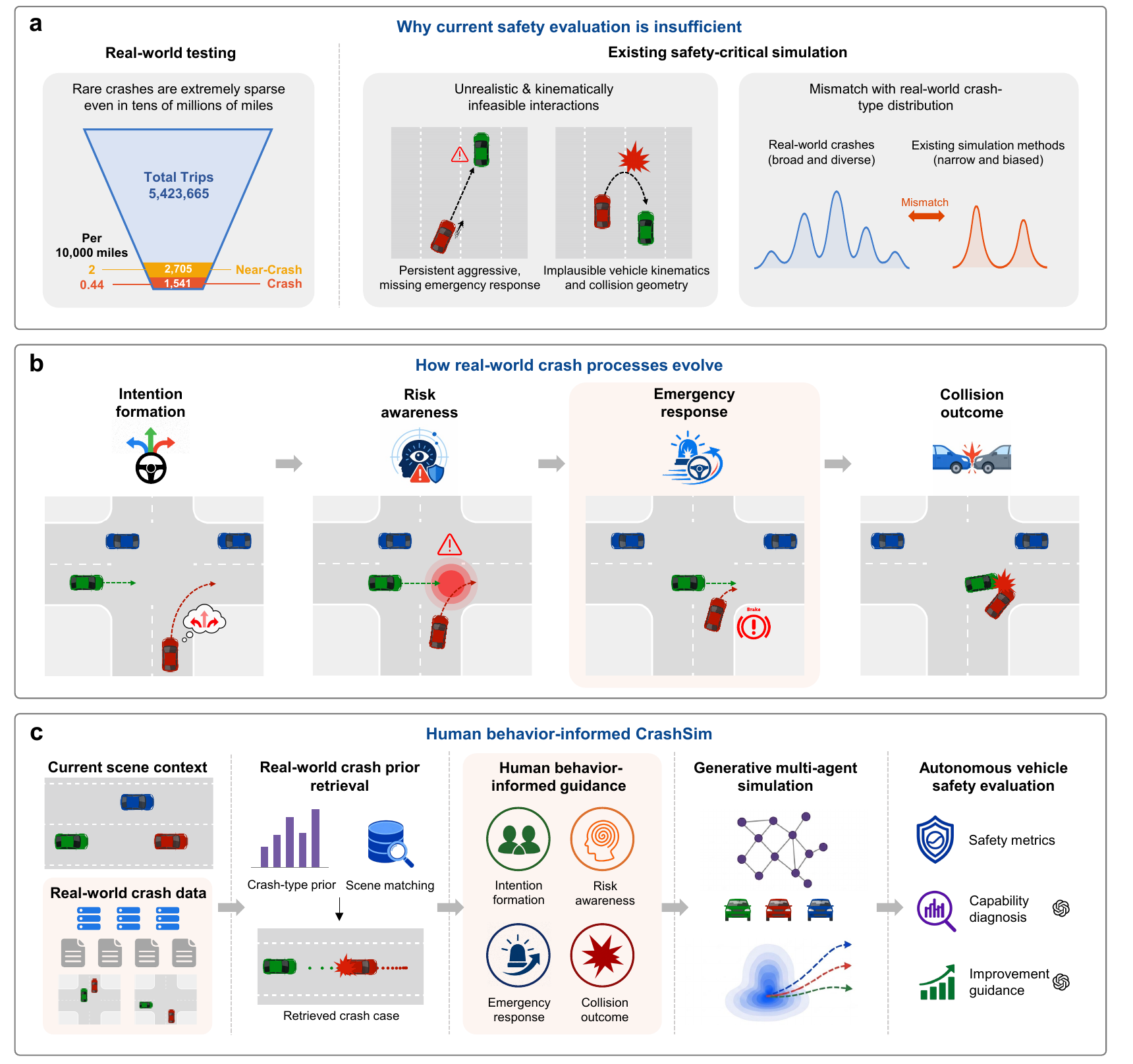}
\caption{\textbf{Challenges in autonomous vehicle safety evaluation and overview of CrashSim.} \textbf{a} Limitations of current autonomous vehicle safety evaluation, including the sparsity of real-world crashes and the behavioral, kinematic, and distributional biases of existing safety-critical simulation. Statistics in the real-world testing funnel are derived from ref.~\cite{njord2015bigdata}. \textbf{b} Real-world crash processes evolve from intention formation through risk awareness and emergency response to the eventual collision outcome whereas existing safety-critical simulation often overlooks the emergency response and leads to aggressive collisions. \textbf{c} CrashSim retrieves real-world crash priors based on the current scene context and uses them to provide human behavior-informed guidance for generative multi-agent simulation and autonomous vehicle safety evaluation.}
\label{fig:introduction}
\end{figure}

Road traffic safety remains a fundamental societal challenge. As autonomous driving rapidly develops and becomes increasingly integrated into modern transportation systems, reliable safety evaluation of autonomous vehicles is essential not only for improving road safety, but also for supporting the broader deployment and continued development of autonomous driving technologies \cite{tafidis2022safety,sohrabi2021quantifying,nunes2018people}. However, safety-critical events, particularly crashes and severe near-crashes, are extremely rare in real-world driving. Reliable assessment of vehicle safety requires sufficient data on such events, but they account for only a very small proportion of real-world driving data \cite{liu2024curse, njord2015bigdata}, even in datasets comprising millions of trips or tens of millions of miles of driving, as illustrated in Fig.~\ref{fig:introduction}a. Consequently, adequately assessing vehicle safety through real-world road testing alone can require a very large amount of driving, making the evaluation process both time-consuming and costly \cite{Kalra2016Driving}. To enable more efficient evaluation of rare and hazardous interactions, safety-critical scenario generation in simulation has become an important component of autonomous vehicle safety evaluation \cite{ding2023survey}, and maintaining realism in both driving behavior and scene-level distributions is critical to reliable safety evaluation.

Existing studies have developed diverse approaches for exposing autonomous vehicles to safety-critical interactions in simulation. Environment-based testing promotes the occurrence of rare and safety-critical events through targeted scenario design or accelerated sampling \cite{dosovitskiy2017carla,lopez2018microscopic, Feng2021,feng2023dense,qian2026test, Yan2023}; optimization-based methods optimize scenario parameters or agent behaviours towards more critical or collision-inducing interactions \cite{wang2021advsim, Rempe2022,hanselmann2022king}; generative approaches combine learned naturalistic traffic distributions with guidance towards safety-critical outcomes \cite{chang2024safe, Peng2025GuidedLDM, Peng2026LDScene,wu2026vlm}; and recent LLM-based methods directly exploit the reasoning capabilities of LLMs to generate collision trajectories \cite{mei2025llm,jiang2026anchordrive,chang2024llmscenario}. Together, these advances have substantially improved the efficiency of safety-critical testing. However, existing methods commonly prioritize increasing collision occurrence or scenario criticality, while giving limited attention to modelling the interaction process preceding impact. This limitation is particularly important because real-world crashes typically emerge through evolving interactions rather than as isolated collision outcomes. As a result, existing methods may generate crashes characterized by persistent aggressive motion, missing emergency responses or implausible vehicle kinematics. Moreover, existing approaches rarely explicitly account for the scene-level distribution of generated crashes, which can cause generated scenario sets to over-represent particular conflict types and deviate from the composition of real-world crashes. As illustrated in Fig.~\ref{fig:introduction}a, these limitations highlight two dimensions of realism in crash scenario generation: the pre-collision interaction process within individual scenarios and the crash-type distribution across the generated scenario set.

Real-world crash data provide a basis for addressing the two dimensions of realism discussed above. At the level of a single crash scenario, crashes typically develop through a temporally evolving process rather than occurring as isolated terminal events. Analyses of real crashes and near-crashes indicate that drivers exhibit diverse braking and evasive responses before a collision \cite{svard2021computational,guo2023modeling}. Models of human collision avoidance further characterize this process as a dynamic interplay between threat detection, response selection, and evasive action execution \cite{schumann2026active}. To characterize this temporal evolution for crash simulation, we represent crash development through four stages: intention formation, risk awareness, emergency response, and collision outcome, as illustrated in Fig.~\ref{fig:introduction}b. At the level of the scenario set, real crashes also exhibit a structured distribution across interaction types. Based on real-world crash databases, the National Highway Traffic Safety Administration (NHTSA) developed a pre-crash typology that organizes collision-producing interactions according to their preceding manoeuvres and conflict configurations \cite{najm2007precrash}. The resulting statistics reveal a highly non-uniform distribution across crash types, with longitudinal conflicts, lane-change and lateral interactions, intersection encounters, and opposing-direction conflicts occurring at markedly different frequencies. Real-world crash data thus serve as a valuable source of information for guiding the construction of realistic crash simulations.

However, learning such real-world crash characteristics directly from crash data remains challenging. In contrast to naturalistic driving datasets, which provide abundant and relatively standardized trajectories for training generative traffic models, real-world crash data are sparse, unevenly distributed across interaction types, and heterogeneous in representation. Such heterogeneous crash data span structured crash reports and taxonomies \cite{najm2007precrash}, naturalistic crash and near-crash records \cite{Hankey2016SHRP2}, reconstructed trajectories \cite{Jiao2026GSSM}, and crash video recordings \cite{fang2021dada,yao2022dota}. Only a limited fraction of these data provide sufficiently complete and consistent trajectories for direct generative modelling, making it difficult to train a general-purpose crash generator that captures the diversity of real-world collision processes and crash types. Retrieval-augmented generation (RAG) offers a practical strategy for leveraging such sparse and heterogeneous crash data by retrieving contextually relevant information at inference time, without requiring all available crash information to be represented in a homogeneous training corpus \cite{Lewis2020RAG}. Meanwhile, large language models (LLMs) and vision-language models (VLMs) have demonstrated strong capabilities for reasoning over complex scene context and heterogeneous forms of prior knowledge \cite{shi2025evidence,qiu2025quantifying}, and have increasingly been applied to safety-critical driving scenario generation \cite{mei2025llm,jiang2026anchordrive,chang2024llmscenario}. Inspired by these developments, we leverage retrieval-based crash priors together with the semantic reasoning capabilities of LLMs and VLMs to incorporate heterogeneous real-world crash data into generative modelling without relying on large-scale, uniformly structured crash trajectories.

Here we present CrashSim, a novel framework that incorporates real-world crash priors into generative multi-agent traffic simulation (Fig.~\ref{fig:introduction}c). Given a naturalistic driving scene, the VLM agent interprets the road context and traffic interactions among nearby road users, together with real-world crash-type statistics, to identify a potential safety-critical interaction. CrashSim then retrieves a relevant real-world crash prior through hierarchical matching of interaction type, event context, and temporal motion pattern. The retrieved crash prior is transformed into human behavior-informed guidance that preserves its relative-motion pattern and behavior-stage timing while adapting them to the geometry and interaction context of the current scene. This guidance steers the diffusion-based generator, pretrained on naturalistic driving data, towards realistic safety-critical interactions. Because real-world crash data are inherently sparse, CrashSim organizes the available cases into a small-scale crash retrieval library rather than relying on a large crash trajectory dataset for model training. By transferring the crash priors contained in this library to diverse naturalistic driving scenes, CrashSim can generate a substantially larger set of crash scenarios while preserving both realistic temporal evolution at the individual-scenario level and the real-world distribution of crash types across the generated scenario set.

We show that CrashSim improves realism at both the individual-scenario and scenario-set levels, more closely reproducing real-world pre-impact behavior and collision characteristics while also matching the observed distribution of crash types. We further use CrashSim to construct nuCrash, a dataset of more than 4,000 crash and near-crash scenarios that enriches rare safety-critical interactions while retaining these behavioral and distributional characteristics. Closed-loop evaluation of five autonomous-driving planners shows that nuCrash more clearly differentiates their safety performance than naturalistic nuScenes. An LLM-assisted evaluation agent further relates these differences to planner-specific capability limitations and failure patterns and provides targeted improvement guidance, thereby supporting interpretable safety evaluation. In contrast to existing autonomous-driving evaluation frameworks that primarily rely on predefined quantitative metrics \cite{caesar2021nuplan,peng2026nuplan}, our approach provides a capability-oriented interpretation of planner performance under safety-critical interactions. Together, these results demonstrate that sparse real-world crash priors can be transferred across naturalistic driving contexts to improve the realism of safety-critical scenario generation, while LLM-assisted analysis enables more informative and interpretable autonomous-vehicle safety evaluation.

\section*{Results}
\label{sec2}
This study models real-world crashes as temporally evolving processes comprising four stages: intention formation, risk awareness, emergency response and collision outcome, rather than treating collision occurrence as a single generation objective. We first show that CrashSim more faithfully reproduces the temporal progression, pre-impact dynamics, response timing and collision behaviors observed in real-world crashes, while also better preserving real-world crash-type distributions than competing scenario-generation methods. We then use CrashSim to construct nuCrash, a long-tail safety-critical dataset that substantially increases the prevalence and diversity of crash and near-crash scenarios while retaining key behavioral and distributional characteristics of real-world crashes. Closed-loop evaluation of five autonomous driving planners shows that nuCrash reveals substantially stronger planner-specific performance differences than naturalistic nuScenes, and that the LLM-based evaluation agent can translate deterministic metrics and scenario-level evidence into capability assessments, failure diagnoses and actionable improvement guidance. Finally, ablation studies show that real-world crash prior retrieval primarily improves the fidelity of collision-process dynamics, and VLM-based scene understanding improves the alignment of generated crash-type distributions with real-world data.

\subsection*{Experimental setup}

\paragraph{Dataset.} CrashSim generates realistic crash scenarios from naturalistic driving contexts under the guidance of real-world crash priors. We use nuScenes \cite{nuscenes2019} as the source of naturalistic initial scenes. The dataset contains 1,000 urban driving scenes of 20\,s each, from which we extract the motion trajectories of the ego vehicle and surrounding traffic participants together with the associated road context. To characterize how real-world crashes evolve and how different crash types are distributed, we incorporate two sources of crash prior knowledge \cite{Jiao2026GSSM,najm2007precrash}. We use reconstructed crash trajectories \cite{Jiao2026GSSM}, derived from the SHRP2 NDS \cite{antin2019second,Hankey2016SHRP2}, to characterize the pre-collision evolution and vehicle dynamics of real-world crashes. The NHTSA pre-crash scenario typology \cite{najm2007precrash} provides complementary empirical statistics on the distribution of real-world crash types. Using these naturalistic driving contexts and real-world crash priors, we construct the nuCrash dataset with CrashSim. nuCrash comprises over 4,000 safety-critical scenarios, including crashes and near-crashes across diverse longitudinal and lateral interactions. Its crash-category distribution broadly follows that observed in real-world crash data and spans diverse long-tail collision scenarios, enabling safety evaluation across a wide range of crash types. We further compare nuCrash with SimEngine \cite{li2026world}, an existing long-tail driving dataset.

\paragraph{Scenario-generation baselines.} We compare CrashSim with two safety-critical scenario-generation methods. Strive \cite{Rempe2022} uses a graph-based variational autoencoder to learn a traffic-motion prior and generates collision scenarios through latent-space adversarial optimization, whereas LD-scene \cite{Peng2026LDScene} uses LLM-generated guidance to steer a diffusion model towards adversarial interactions. Both methods primarily optimize towards collision outcomes rather than modeling how real-world crashes evolve through a temporal collision process, which can result in overly aggressive and behaviorally implausible collisions.

\paragraph{Evaluation metrics.} We evaluate collision scenario realism from three aspects: collision-process fidelity, behavioral plausibility and scene-level distributional realism. Collision-process fidelity is assessed using the temporal composition of intention formation, risk awareness, emergency response and collision, together with the temporal evolution of risk awareness and emergency response, pre-impact longitudinal acceleration, reaction time, response time and braking severity. Behavioral plausibility is evaluated from collision impact locations and the proportions of adversarial and other non-ego vehicles leaving the drivable area. Scene-level distributional realism is assessed across sixteen fine-grained crash types and five high-level conflict categories, with Total Variation (TV) distance \cite{gibbs2002choosing} and Jensen-Shannon divergence (JSD) \cite{lin1991divergence} quantifying discrepancies from real-world distributions. For dataset-level comparison, we additionally report crash and near-crash rates to characterize the prevalence of safety-critical events.

\paragraph{Planner evaluation.} We evaluate five autonomous driving planners spanning rule-based and learning-based approaches: Lane-graph planner \cite{montemerlo2008junior}, PDM-closed planner \cite{dauner2023parting}, imitation learning-based (IL) planner \cite{bansal2018chauffeurnet}, imitation learning-based multi-agent motion (IL-Multi) planner \cite{suo2021trafficsim} and Diffusion planner \cite{ICLR2025_5c0cd430}. Each planner is evaluated in closed loop on both naturalistic nuScenes scenarios and safety-critical nuCrash scenarios. Following the temporal setting of the nuScenes motion planning task, at each replanning step, the planner uses 2 seconds (4 steps) of motion history to predict a 6 seconds (12 steps) future trajectory, with replanning performed every 0.5\,s based on the updated scene state. An LLM-based evaluation agent further analyzes deterministic metrics and scenario-level evidence across the two datasets to assess planner capabilities, diagnose failure mechanisms, and identify directions for planner improvement.

\begin{figure}
    \centering
    \includegraphics[trim={15 315 15 15}, clip, width=\linewidth]{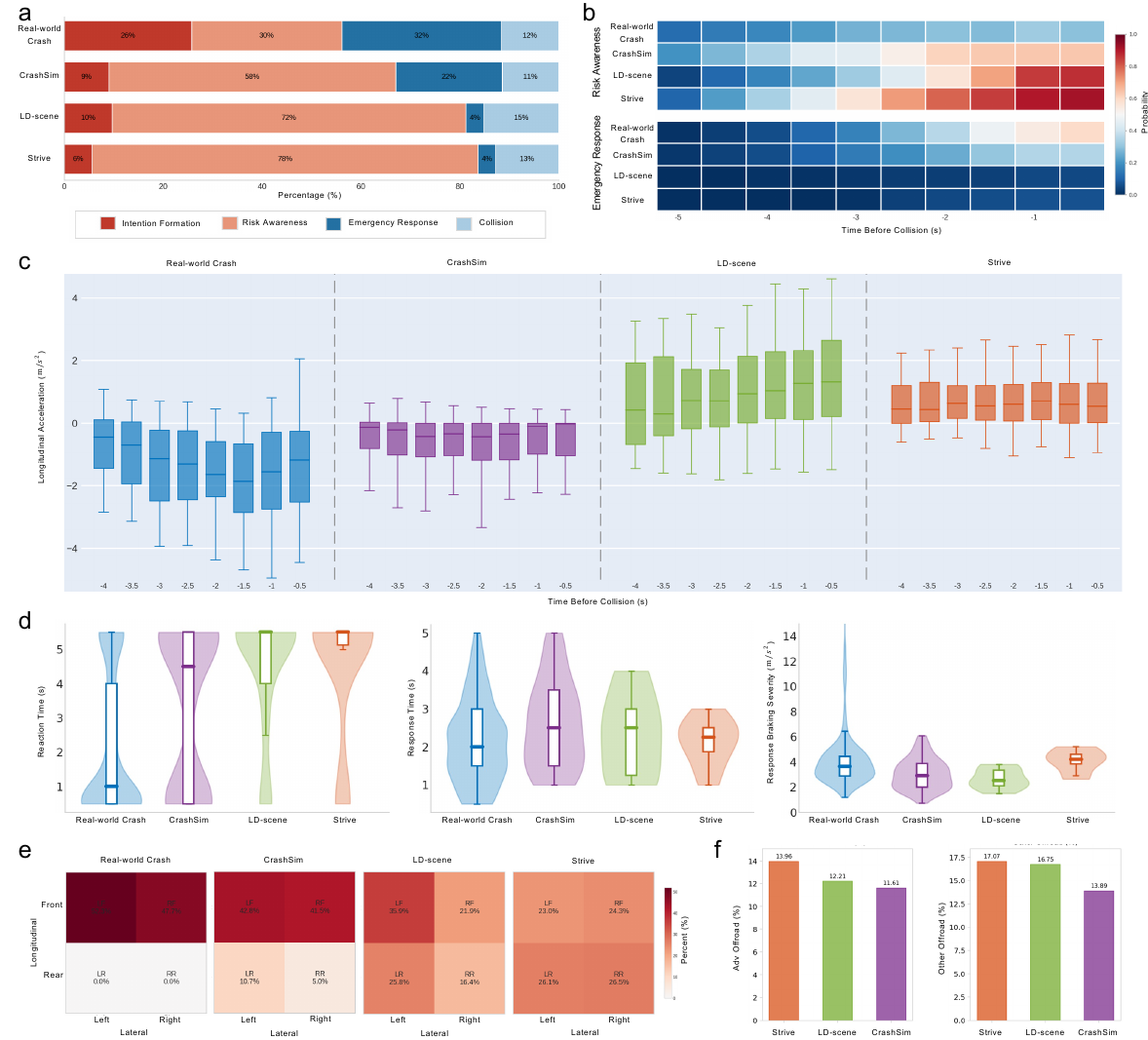}
    \caption{
        \textbf{CrashSim generates realistic human behavior-informed collision processes.}
        \textbf{a} Temporal composition of intention formation, risk awareness, emergency response and collision in real-world crashes and generated scenarios. CrashSim retains the complete four-stage progression and more closely reproduces the stage proportions observed in real-world crashes.
        \textbf{b} Time-resolved probability heatmaps showing the evolution of risk awareness and emergency response before collision. CrashSim captures the gradual development of risk awareness followed by the emergence of emergency response.
        \textbf{c} Temporal distributions of longitudinal acceleration during the final four seconds preceding collision. CrashSim reproduces the predominantly decelerative response and its temporal evolution in real-world crashes.
        \textbf{d} Distributions of reaction time response time and braking severity across real-world crashes and generated scenarios. Together \textbf{a--d} show that CrashSim captures key human behavioral dynamics underlying crash progression including risk development response timing and emergency braking.
        \textbf{e} Spatial distributions of collision impact locations across the front, rear, left and right regions of the ego vehicle. CrashSim generates behaviorally realistic ego-at-fault collisions, characterized by a predominance of frontal impacts, thereby enabling more effective evaluation of ego-planner safety.
        \textbf{f} Percentages of adversarial and other non-ego vehicles that leave the drivable area. The low off-road rates produced by CrashSim indicate physically plausible pre-collision interactions and stronger compliance with road boundaries. Together, \textbf{e-f} demonstrate the behavioral plausibility of CrashSim-generated collision scenarios.
    }
    \label{fig:human_aware}
\end{figure}

\subsection*{CrashSim reproduces realistic human behavior-informed collision processes}
Real-world collision scenarios cannot be characterized by a single driving action, but instead unfold through a temporally evolving sequence of intention formation, risk awareness and emergency response. We therefore investigate whether the generated scenarios reproduce this complete behavioral progression together with the associated driving dynamics and collision outcomes as shown in Fig.~\ref{fig:human_aware}.

CrashSim more faithfully reproduces the temporal progression observed in real-world crashes than competing methods. Fig.~\ref{fig:human_aware}a, b shows that CrashSim preserves the temporal allocation of intention formation, risk awareness, emergency response and collision, while recovering the gradual transition from risk perception to emergency action before impact. In contrast, LD-scene and Strive remain in the risk-awareness stage for prolonged periods and exhibit little emergency response immediately before collision, resulting in overly aggressive and behaviorally implausible collision processes. This difference is also reflected in the longitudinal dynamics shown in Fig.~\ref{fig:human_aware}c. Real-world crashes exhibit increasingly negative longitudinal acceleration as the collision approaches, and CrashSim reproduces this decelerative tendency, whereas LD-scene and Strive maintain predominantly positive acceleration into the conflict with the ego vehicle. Fig.~\ref{fig:human_aware}d further compares the distributions of reaction time and response time, together with braking severity during the emergency-response stage. Across these measures, CrashSim shows closer overall agreement with the real-world distributions in both response timing and braking intensity. Together, these results indicate that CrashSim reproduces the evolving behavioral and dynamic processes preceding a collision, rather than merely producing the terminal collision outcome.

CrashSim also generates collision behaviors that more closely resemble those observed in real-world crashes. Fig.~\ref{fig:human_aware}e shows that the impact-location distribution generated by CrashSim is highly consistent with real-world crash statistics, particularly in reproducing the predominance of frontal impacts characteristic of ego-at-fault collisions, thereby enabling more effective evaluation of ego-planner safety. Fig.~\ref{fig:human_aware}f further shows that adversarial and surrounding vehicles are substantially less likely to contact the curb before collision than in scenarios generated by competing methods. Together, these results demonstrate that CrashSim improves both behavioural realism and kinematic feasibility, yielding more effective safety-critical driving scenarios for evaluating autonomous driving planners.

Collectively these results show that CrashSim reproduces realistic collision processes through temporally coherent human responses realistic driving dynamics and plausible collision outcomes. This enables the generation of safety-critical scenarios with substantially higher behavioral realism than existing scenario generation methods.

\begin{figure}
    \centering
    \includegraphics[trim={45 570 70 18}, clip, width=\linewidth]{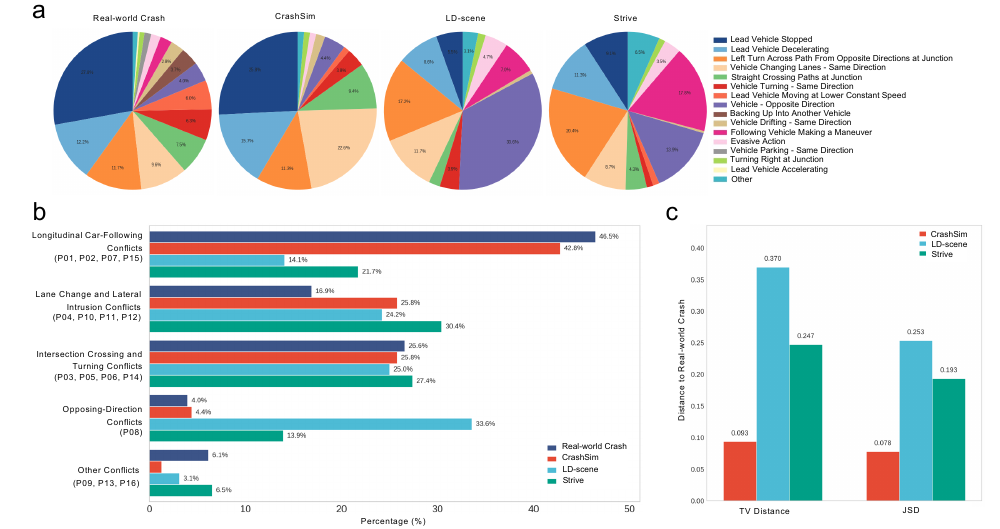}
    \caption{
        \textbf{CrashSim preserves realistic crash scenario distributions.}
        \textbf{a} Distribution of fine-grained crash scenarios across real-world crashes and generated datasets.
        \textbf{b} Distribution of five high-level crash categories, including longitudinal car-following conflicts, lane-change and lateral intrusion conflicts, intersection crossing and turning conflicts, opposing-direction conflicts, and other conflicts.
        \textbf{c} Distributional discrepancy from real-world crashes, quantified using total variation distance for the five high-level categories and Jensen-Shannon divergence for the sixteen fine-grained crash types.
        Across both levels of semantic granularity, CrashSim most closely reproduces real-world crash statistics and achieves the lowest distributional discrepancy among the compared methods.
    }
    \label{fig:scenario_distribution}
\end{figure}

\subsection*{CrashSim preserves realistic crash scenario distributions}

Generating realistic individual crashes does not necessarily imply that the resulting scenario set reflects the diversity of real-world traffic conflicts. We therefore evaluate whether CrashSim reproduces the distribution of crash scenarios observed in naturalistic crash datasets across both fine-grained crash types and higher-level conflict categories, as summarized in Fig.~\ref{fig:scenario_distribution}.

CrashSim closely reproduces the distribution of fine-grained crash scenarios observed in real-world data. As shown in Fig.~\ref{fig:scenario_distribution}a, the proportions of sixteen representative crash types generated by CrashSim align well with those in the real-world dataset, whereas LD-scene and Strive deviate substantially across multiple scenario categories. The relatively high proportion of Vehicle Changing Lanes scenarios reflects the scene-conditioned generation process of CrashSim. Although the real-world crash-type distribution is incorporated as an auxiliary prior during scenario reasoning, the identification of potential safety-critical interactions remains primarily grounded in the current scene context. Since the original nuScenes scenes contain a relatively large number of interactions between vehicles in adjacent lanes, lane-change-related conflicts are correspondingly more frequent in the generated scenarios. At the aggregated level, Fig.~\ref{fig:scenario_distribution}b shows that CrashSim most faithfully preserves the relative frequencies of longitudinal car-following, lane-change, intersection, opposing-direction and other conflicts. Together, these results demonstrate that CrashSim can generate safety-critical scenarios across diverse driving contexts while remaining aligned with real-world crash-type distributions.

We further quantify distributional realism using TV distance and JSD between generated and real-world crash distributions. Fig.~\ref{fig:scenario_distribution}c shows that CrashSim consistently achieves the smallest distributional discrepancy under both metrics, substantially outperforming LD-scene and Strive. The improvement under two distributional measures demonstrates that CrashSim more faithfully preserves the overall statistical composition of real-world crash scenarios.

Collectively, these results demonstrate that CrashSim not only generates realistic individual collision scenarios but also faithfully reconstructs the population-level distribution of real-world crashes. This capability enables more representative, diverse and statistically realistic safety-critical scenario generation for autonomous vehicle evaluation.

\begin{figure}
    \centering
    \includegraphics[trim={35 545 50 22}, clip, width=\linewidth]{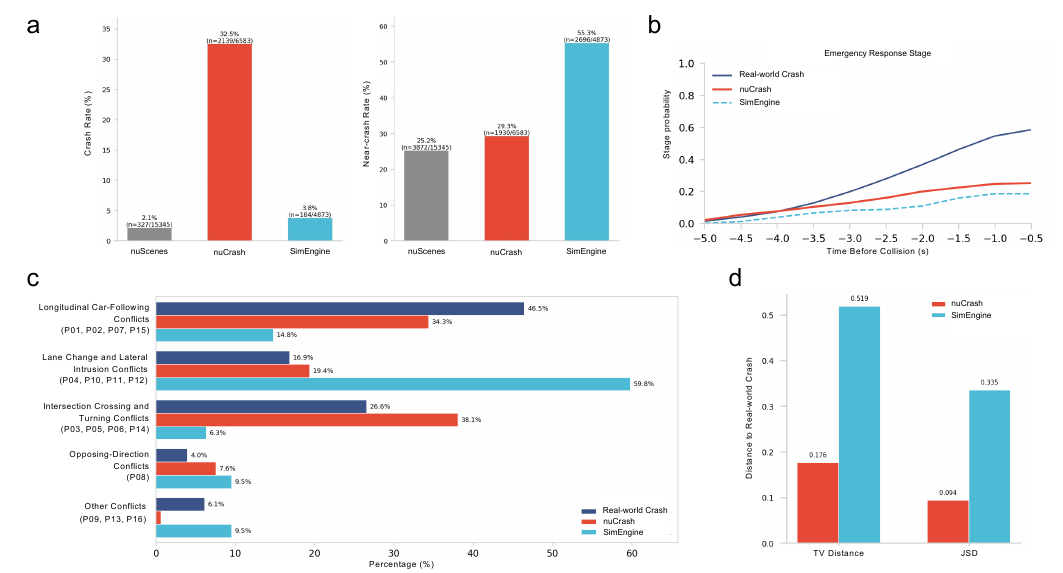}
    \caption{
    \textbf{nuCrash constructs a realistic long-tail crash dataset.}
    \textbf{a} Comparison of crash rate and near-crash rate among nuScenes, nuCrash, and SimEngine.
    \textbf{b} Temporal evolution of the emergency response stage before collision in real-world crashes, nuCrash, and SimEngine.
    \textbf{c} Distribution of high-level crash categories in real-world crashes, nuCrash, and SimEngine.
    \textbf{d} Distributional discrepancy from real-world crashes, measured using Total Variation distance and Jensen-Shannon divergence.
    \textbf{a} shows that nuCrash substantially increases the prevalence of realized collisions relative to the naturalistic nuScenes dataset, whereas SimEngine is dominated by near-crash cases and contains relatively few actual collisions. \textbf{b-d} demonstrate that nuCrash more faithfully reproduces the temporal dynamics and scenario distributions observed in real-world crashes than the existing long-tail dataset. Collectively, these results show that nuCrash provides a more realistic and challenging dataset for autonomous vehicle safety evaluation.
    }
    \label{fig:nucrash}
\end{figure}

\begin{figure}
    \centering
    \includegraphics[trim={30 388 25 15}, clip, width=\linewidth]{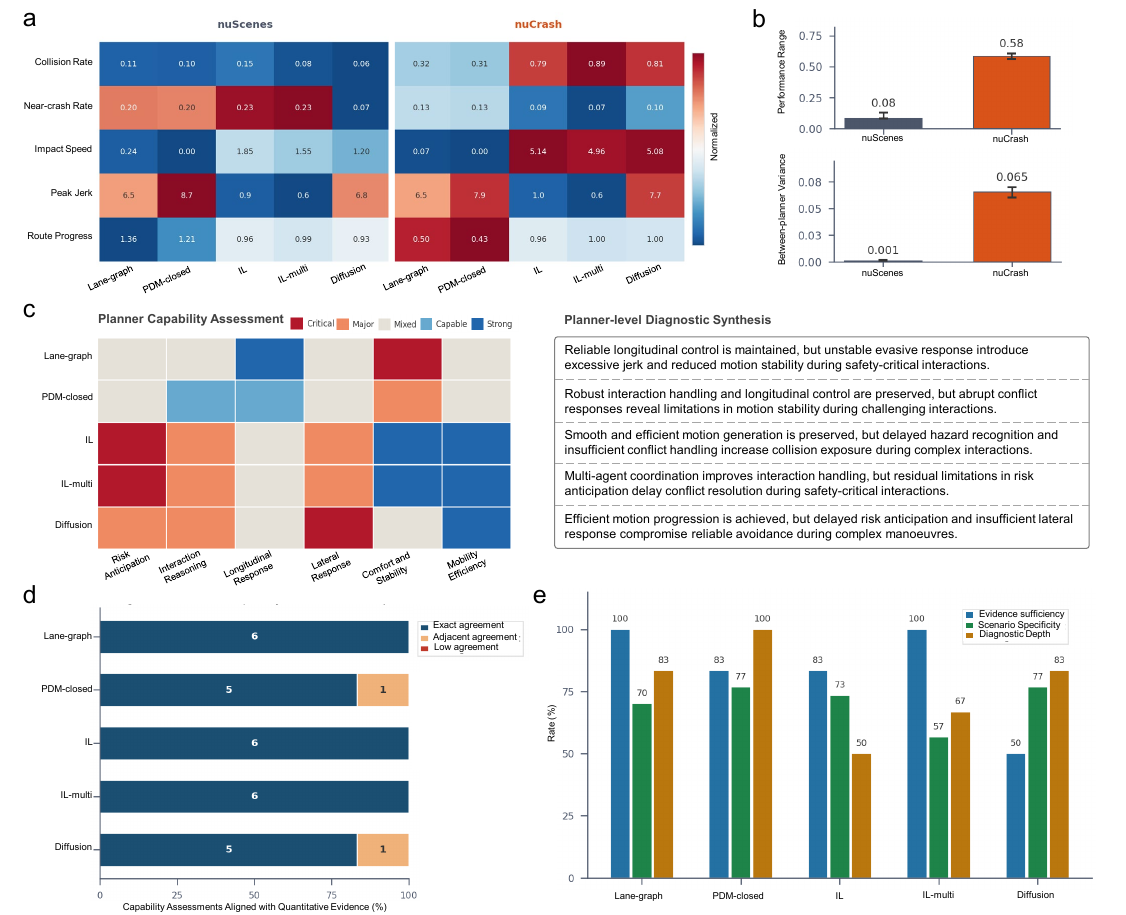}
    \caption{
    \textbf{Interpretable and reliable LLM-based evaluation of autonomous driving planners.}
    \textbf{a} Planner performance across five safety and driving-quality metrics on nuScenes and nuCrash. Values are normalized within each metric for visual comparison.
    \textbf{b} Performance range and between-planner variance on the two datasets, showing substantially stronger planner differentiation on nuCrash.
    \textbf{c} LLM-based capability assessment across six planner capabilities together with planner-level diagnostic synthesis that translates quantitative evaluation evidence into interpretable planner strengths and limitations.
    \textbf{d} Alignment between capability assessments and quantitative evidence.
    \textbf{e} Planner-level audit results quantifying the evidence sufficiency, scenario specificity, and diagnostic depth. Together these analyses demonstrate that the LLM-based evaluation agent can provide interpretable and reliable planner assessments.
    }
    \label{fig:safety}
\end{figure}

\subsection*{nuCrash provides a realistic long-tail crash dataset}

Naturalistic driving datasets, such as nuScenes, faithfully capture everyday traffic interactions but contain relatively few safety-critical events, limiting their coverage of the long-tail conditions encountered in real-world autonomous driving. Using scenes from nuScenes as initial scenarios and CrashSim as the scenario generation framework, we construct nuCrash, a large-scale safety-critical dataset comprising diverse crash and near-crash scenarios. As shown in Fig.~\ref{fig:nucrash}, we compare nuCrash with both the original nuScenes dataset and SimEngine, an existing synthetic long-tail dataset.

Compared with both nuScenes and SimEngine, nuCrash contains a substantially larger proportion of safety-critical scenarios, making it a more challenging long-tail crash dataset. As shown in Fig.~\ref{fig:nucrash}a, nuCrash achieves the highest crash rate among the three datasets while maintaining a balanced composition of crash and near-crash events. In contrast, nuScenes contains only a limited number of hazardous interactions, whereas SimEngine is dominated by near-crash scenarios. 

Beyond increasing scenario criticality, nuCrash also preserves the behavioral and statistical characteristics of real-world crashes. As shown in Fig.~\ref{fig:nucrash}b, the temporal evolution of emergency responses in nuCrash closely tracks that observed in the real-world crash dataset, whereas SimEngine exhibits markedly different response dynamics. Fig.~\ref{fig:nucrash}c further shows that nuCrash reproduces the distribution of high-level crash categories across longitudinal, lane-change, intersection, opposing-direction, and other conflicts. In contrast, SimEngine substantially overrepresents lane-change conflicts while underrepresenting several other categories. These observations are quantitatively confirmed by the TV distance and JSD to the real-world crash distribution shown in Fig.~\ref{fig:nucrash}d. nuCrash achieves consistently smaller distances than SimEngine under both metrics, demonstrating that it enriches safety-critical events without compromising the behavioral and distributional realism of real-world crashes.

Collectively, these results demonstrate that nuCrash combines the realism of naturalistic crash data with substantially richer safety-critical interactions. By preserving both the behavioral dynamics and statistical characteristics of real-world crashes while significantly increasing the prevalence of hazardous scenarios, nuCrash provides a realistic and challenging long-tail crash dataset for autonomous vehicle safety evaluation.

\begin{figure}
    \centering
    \includegraphics[trim={60 570 62 15}, clip, width=\linewidth]{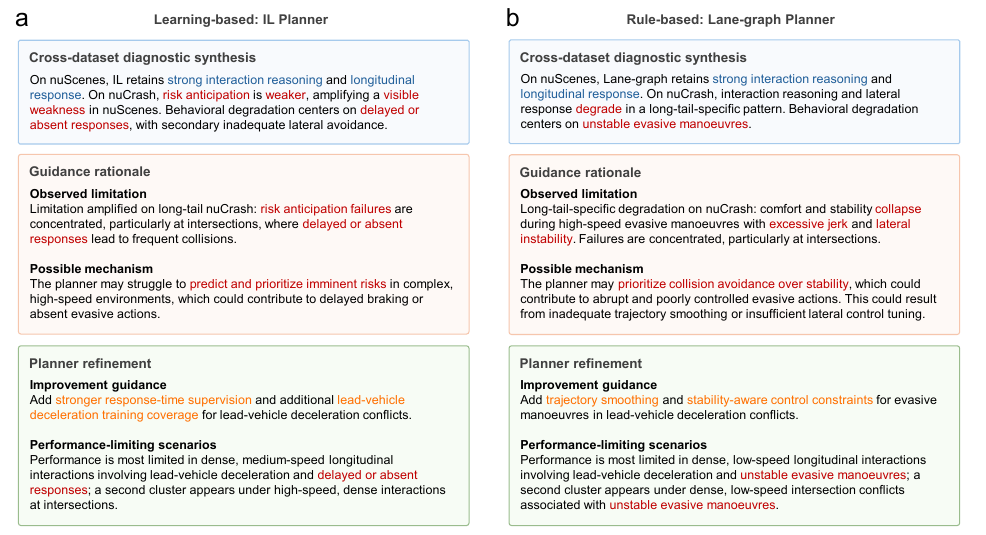}
    \caption{
    \textbf{Qualitative case studies of LLM-based planner evaluation.}
    \textbf{a} Learning-based IL Planner.
    \textbf{b} Rule-based Lane-graph Planner. The LLM-based evaluation agent synthesizes cross-dataset planner behavior across naturalistic and long-tail scenarios. It identifies performance degradation and possible failure mechanisms. It further translates the diagnostic findings into planner refinement guidance and highlights representative performance-limiting scenarios. These cases illustrate the ability of the LLM-based evaluation to provide interpretable and actionable planner assessments across distinct planner paradigms.
    }
    \label{fig:case}
\end{figure}

\subsection*{CrashSim enables interpretable autonomous vehicle safety evaluation}
Beyond generating safety-critical scenarios, CrashSim supports interpretable safety evaluation by combining closed-loop planner performance with LLM-based diagnostic analysis. Fig.~\ref{fig:safety}a shows clear differences in planner behavior between naturalistic and long-tail scenarios. These differences become substantially more pronounced on nuCrash. Fig.~\ref{fig:safety}b quantifies this effect. The normalized performance range increases from 0.08 on nuScenes to 0.58 on nuCrash while the between-planner variance rises from 0.001 to 0.065. The increased separation reflects distinct planner-specific weaknesses under safety-critical interactions rather than uniform degradation. Learning-based planners mainly exhibit increased collision severity while rule-based planners show more pronounced degradation in motion stability or route completion.

Fig.~\ref{fig:safety}c shows that the LLM-based evaluation agent translates this quantitative evidence into capability assessments and planner-level diagnostic summaries. For example, as shown in Fig.~\ref{fig:safety}c, the IL planner is assessed as having strong comfort and stability but critical limitations in risk anticipation. This assessment is consistent with the quantitative results in Fig.~\ref{fig:safety}a where IL exhibits low peak jerk but substantially increased crash rate on nuCrash.
The diagnostic summaries further associate these capability limitations with concrete behavioral patterns such as delayed hazard recognition or unstable evasive responses. The resulting assessments therefore characterize not only whether planner performance deteriorates but also which capabilities are most affected under long-tail conditions.

The reliability of these assessments is examined in Fig.~\ref{fig:safety}d and Fig.~\ref{fig:safety}e. Fig.~\ref{fig:safety}d shows that the capability assessments are highly consistent with the corresponding quantitative evidence across planners. The planner-level audit in Fig.~\ref{fig:safety}e further shows generally high evidence sufficiency and diagnostic depth while revealing some variation in scenario specificity across planners. These results indicate that the LLM-based evaluation remains closely grounded in quantitative evidence while providing planner-specific and interpretable diagnostic insights.

Fig.~\ref{fig:case} presents case studies of the LLM-based planner evaluation for representative learning-based and rule-based planners. For IL planner, the dominant degradation on nuCrash is weakened risk anticipation with delayed or absent responses emerging as a major failure pattern particularly in rear-end collision scenarios triggered by lead-vehicle deceleration. The evaluation agent further attributes this vulnerability to insufficient exposure of the learning-based planner to long-tail collision patterns during training and consequently recommends stronger response-time supervision and additional training coverage for lead-vehicle deceleration conflicts. The Lane-graph planner exhibits a different limitation with long-tail degradation dominated by unstable evasive manoeuvres and excessive jerk. The corresponding guidance emphasizes trajectory smoothing and stability-aware control constraints. The identified performance-limiting scenarios further enable targeted validation of whether the proposed improvement guidance effectively mitigates the diagnosed planner weaknesses. Together, these cases demonstrate that CrashSim extends safety evaluation beyond performance measurement by connecting long-tail failures with capability limitations and actionable directions for planner refinement.

\begin{figure}
    \centering
    \includegraphics[trim={35 612 30 20}, clip, width=\linewidth]{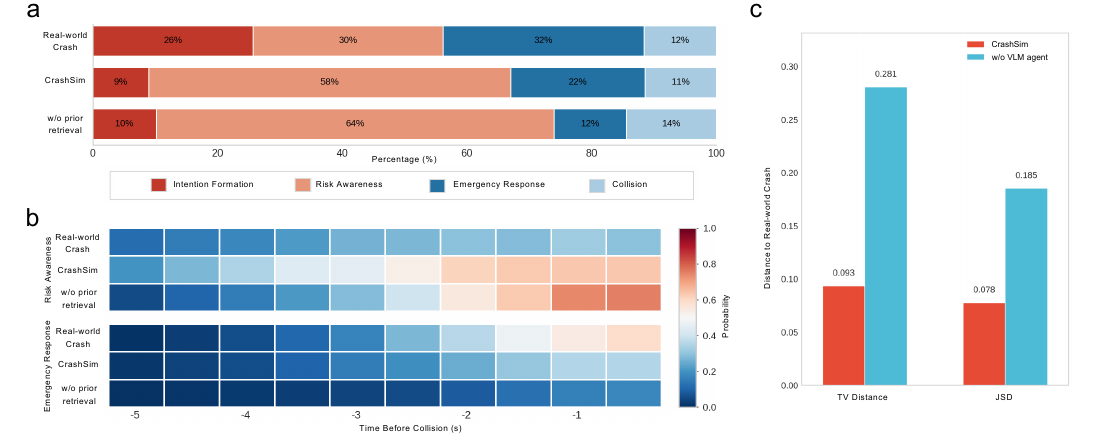}
    \caption{
        \textbf{Ablation studies of the crash-prior retrieval module and the VLM agent.}
        \textbf{a} Temporal composition of the collision process in real-world crashes, CrashSim and CrashSim without the crash-prior retrieval module. Removing the crash-prior retrieval module increases the discrepancy from the real-world composition, indicating that retrieved crash knowledge helps preserve realistic collision-process characteristics.
        \textbf{b} Temporal evolution of risk awareness and emergency response probabilities for the same settings. The complete framework more closely reproduces the temporal progression observed in real-world crashes, demonstrating that the crash-prior retrieval module improves the realism of pre-collision behavioral dynamics.
        \textbf{c} Total variation and Jensen-Shannon divergence between generated and real-world crash distributions for CrashSim and CrashSim without the VLM-based scene-understanding agent. The complete framework achieves consistently lower distances, indicating that incorporating real-world crash-type distribution priors into scene-level reasoning improves the alignment of generated scenario types with real-world crash statistics.
    }
    \label{fig:ablation}
\end{figure}

\subsection*{Ablation studies}

We next examined the respective contributions of the real-world crash-prior retrieval module and the VLM-based scene-understanding agent. The crash-prior retrieval module retrieves behavioral and kinematic evidence from real crash cases. The VLM agent interprets the current driving context and identifies the most plausible adversarial vehicle and its potential behavior.

As shown in Fig.~\ref{fig:ablation}a, removing the crash-prior retrieval module reduced the proportion of emergency response from 22\% to 12\%, increasing its discrepancy from the real-world value of 32\%. Figure~\ref{fig:ablation}b further shows that, without crash-prior retrieval, risk awareness developed later and emergency response remained consistently weak. Incorporating retrieved crash evidence promoted earlier risk recognition and a progressively stronger emergency response as the collision approached. These results indicate that the crash-prior retrieval module enables CrashSim to reproduce collision processes that more closely resemble those observed in real-world crashes.

The VLM agent primarily improved the distributional realism across crash scenario types. Fig.~\ref{fig:ablation}c shows that removing this agent increased the TV distance from 0.093 to 0.281 and the JSD from 0.078 to 0.185. The VLM-based agent combines prompts informed by real-world prior distributions with its visual understanding capabilities to analyze the driving scene and identify potential collision patterns consistent with the current driving context. This scene-aware identification improves the agreement between the generated and real-world distributions of crash scenario types.

These ablations reveal the functions of the two components. The crash-prior retrieval module improves the realism of collision-process dynamics, whereas the VLM agent enhances the distributional alignment of crash scenario types. Together, they enable CrashSim to generate crashes that more closely resemble real-world crashes in both collision-process characteristics and scenario-type distributions.

\section*{Discussion}
\label{sec3}
CrashSim addresses a central mismatch between safety-critical traffic simulation and the way crashes develop in real-world driving. Existing generators are effective at increasing the frequency of safety-critical outcomes, yet the final outcome does not ensure that the preceding interaction follows a behaviorally plausible progression. By grounding scenario generation in retrieved real-world crash priors, CrashSim shifts the focus from inducing collisions to reproducing behaviorally plausible pre-collision interaction dynamics. The observed agreement in behavioral-stage composition, pre-impact longitudinal dynamics, response timing, impact location and scene-level crash-type distributions indicates that this grounding improves the realism of both individual interactions and the scene-level distribution of crash types.

CrashSim also produces two resources with broader utility: a retrieval database constructed from real-world crash data, which can provide empirical crash priors for other diffusion-based or LLM-based scenario generation methods, and the nuCrash dataset, which extends naturalistic driving evaluation to safety-critical conditions that are rarely represented in conventional driving datasets. nuScenes provides rich contextual observations of routine traffic, whereas nuCrash systematically enriches safety-critical interactions while preserving realistic behavioral characteristics and introducing diverse crash scenarios whose composition reflects real-world crash distributions. The substantially larger between-planner variation observed on nuCrash suggests that it provides a more discriminative setting for safety evaluation than naturalistic driving data alone. In particular, performance differences that remain weakly expressed during predominantly routine driving become more distinguishable when planners are exposed to rapidly evolving hazardous interactions that demand timely and coordinated responses. This increased sensitivity to planner-specific weaknesses highlights the value of nuCrash for evaluating safety performance under challenging interactions that are underrepresented in naturalistic driving data.

CrashSim further enables a capability-oriented interpretation of closed-loop planner evaluation. Rather than treating the LLM-based evaluation agent as an alternative to quantitative assessment, we use it to jointly analyze deterministic trajectory metrics, scenario-level statistics, failure patterns and their interaction contexts. This analysis translates heterogeneous measurements into structured assessments of planner capabilities and supports cross-dataset reasoning about how these capabilities change from naturalistic to safety-critical driving. Such a formulation connects low-level performance measurements with higher-level behavioral interpretation, helping to identify which aspects of planner behavior are most strongly associated with the observed performance degradation and to derive corresponding directions for planner improvement. More broadly, this illustrates how language-model-based reasoning can serve as an analytical layer over structured simulation evidence, while leaving the underlying performance measurements grounded in deterministic evaluation.

A remaining limitation lies in the diagnostic depth of the LLM-based evaluation. Although the current analysis can associate performance changes with observed behavioral patterns and scenario conditions, its attribution of the underlying failure mechanisms remains relatively coarse. Future work should proceed in three directions. First, the LLM-based evaluation should be extended with deeper attribution analysis to better identify the mechanisms underlying planner failures and their relationships with specific scenario conditions. Second, CrashSim can be further strengthened by incorporating broader real-world crash databases and more reactive modelling of surrounding traffic. Third, safety-critical datasets such as nuCrash could be explored as training resources for improving planner robustness to rare events, with particular attention to integrating safety-critical scenarios into planner training and adaptation without degrading performance under naturalistic driving conditions.

\section*{Methods}
\label{sec5}
\begin{figure}[t]
\centering
\includegraphics[trim={20 750 60 10}, clip, width=\linewidth]{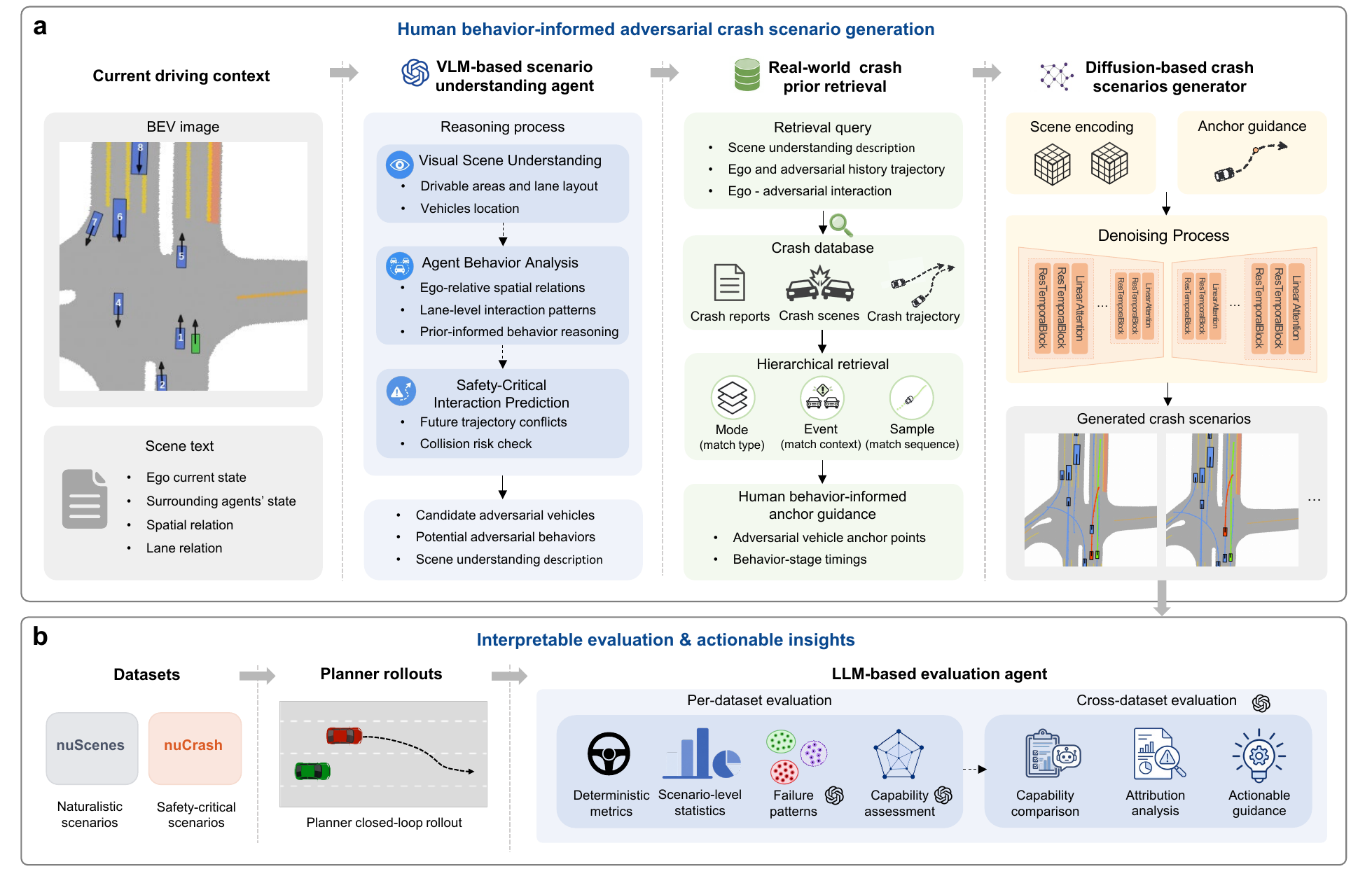}
    \caption{
        \textbf{Human behavior-informed crash generation and interpretable safety evaluation with CrashSim.}
        \textbf{a} Human behavior-informed adversarial crash scenario generation. CrashSim first interprets the current driving context through VLM-based scene understanding and retrieves relevant real-world crash priors through hierarchical matching. The retrieved crash prior provides human behavior-informed anchor guidance for a diffusion-based generator to produce safety-critical crash scenarios.
        \textbf{b} Interpretable evaluation and actionable insights. Generated crash scenarios form the long-tail nuCrash dataset which complements the naturalistic nuScenes dataset for closed-loop planner evaluation. An LLM-based evaluation agent performs both per-dataset and cross-dataset evaluation to diagnose failure mechanisms, assess planner capabilities, and provide actionable guidance for planner improvement. 
    }
    \label{fig:method}
\end{figure}

CrashSim is a human behavior-informed framework for generating realistic safety-critical driving scenarios and evaluating autonomous driving planners under long-tail datasets. The framework combines scene understanding, retrieval of real-world crash priors, and generative multi-agent simulation to introduce realistic crash processes into naturalistic driving contexts. As illustrated in Fig.~\ref{fig:method}a, CrashSim takes a naturalistic driving scene as input, represented by a bird's-eye-view (BEV) image and a structured textual description of the current driving context. A vision language model (VLM)-based scenario understanding agent interprets the scene and identifies candidate vehicles and potential behaviors that may develop into safety-critical interactions. Based on the scene understanding context, the historical trajectories of the ego vehicle and the VLM-identified candidate adversarial vehicle and potential adversarial behavior, and their historical interactions, CrashSim retrieves relevant real-world crash priors through a hierarchical retrieval process. The retrieved prior is then translated into human behavior-informed anchor guidance for a diffusion-based multi-agent trajectory generator, enabling the generation of realistic crash scenarios. This retrieval-augmented generation strategy grounds scenario generation in behavior and motion patterns observed in real-world crashes.

CrashSim further supports interpretable safety evaluation of autonomous driving planners, as shown in Fig.~\ref{fig:method}b. Planners are evaluated in closed loop on both naturalistic nuScenes scenarios and safety-critical nuCrash scenarios. An LLM-based evaluation agent analyzes deterministic metrics and scenario-level evidence across the two datasets to assess planner capabilities, diagnose recurrent failure mechanisms, and provide actionable guidance for planner improvement.

\subsection*{Problem formulation}

We consider safety-critical multi-agent scenario generation for autonomous driving planner evaluation. Each scenario contains $N$ vehicles, comprising an ego vehicle controlled by a planner $\pi$ and $N-1$ surrounding vehicles whose future motions are generated by CrashSim. Among the surrounding vehicles, one is identified as the candidate adversarial vehicle that may develop a safety-critical interaction with the ego vehicle, whereas the others evolve as non-adversarial traffic participants.

A driving scenario is represented by a road map $\mathbf{m}$ and the states of all vehicles. At timestep $t$, the vehicle states are denoted as $\mathbf{s}_{t}=[s^{0}_{t},s^{1}_{t},\ldots,s^{N-1}_{t}]$, where $s^{i}_{t}=(x^{i}_{t},y^{i}_{t},\theta^{i}_{t},v^{i}_{t})$ describes the two-dimensional position, heading and speed of vehicle $i$, and $i=0$ denotes the ego vehicle. The historical trajectories over the previous $T_{\mathrm{hist}}$ timesteps are represented as $\mathbf{x}=\{\mathbf{s}_{t-T_{\mathrm{hist}}},\ldots,\mathbf{s}_{t}\}$. The planner $\pi$ produces the future trajectory of the ego vehicle over a horizon $T$, denoted as $\tau^{0}_{t:t+T}=\pi(\mathbf{m},\mathbf{x})$. 

CrashSim further represents the current driving context using a BEV image $\mathbf{v}$ and a structured textual scene description $\mathbf{n}$. A VLM-based scenario understanding agent processes these inputs as $\mathbf{o}=\phi(\mathbf{v},\mathbf{n})$, where $\mathbf{o}=(i^{\star},b,u)$ specifies the candidate adversarial vehicle $i^{\star}$, its potential adversarial behavior $b$, and a structured textual scene understanding description $u$. Based on this output, CrashSim constructs a structured retrieval query $\mathbf{q}=\left(u,b,\mathbf{r}^{0,i^{\star}},\mathbf{x}^{0},\mathbf{x}^{i^{\star}}\right)$, which integrates the scene understanding, predicted interaction behavior, pairwise interaction information, and historical trajectories of the ego and candidate adversarial vehicles. The retrieval process $\rho$ searches the real-world crash knowledge base $\mathbf{d}_{\mathrm{crash}}$ and returns a matched crash prior, $\mathbf{k}=\rho(\mathbf{q},\mathbf{d}_{\mathrm{crash}})$, which is subsequently transformed into human behavior-informed anchor guidance $\mathbf{a}$ for scenario generation.

Conditioned on the road map $\mathbf{m}$, historical trajectories $\mathbf{x}$ and anchor guidance $\mathbf{a}$, a latent diffusion model $g_{\theta}$ generates the future trajectories of the surrounding vehicles, $\boldsymbol{\tau}_{t+T}=g_{\theta}(\mathbf{m},\mathbf{x},\mathbf{a}).$
The model comprises an encoder $E$ that maps the scene context into a latent representation and a decoder $D$ that reconstructs the generated trajectories from the latent space. For constructing the nuCrash dataset used in this study, the ego vehicle follows its recorded future trajectory in the corresponding nuScenes scene. The resulting scenarios are retained as a fixed evaluation set shared across all planners. During subsequent closed-loop planner evaluation, the recorded ego trajectory is replaced by the trajectory produced by the evaluated planner $\pi$, which is replanned every 0.5 s.

\subsection*{VLM-based scenario understanding agent} 
The VLM-based scenario understanding agent interprets the current driving context and identifies surrounding vehicles that may plausibly develop safety-critical interactions with the ego vehicle. It jointly processes the BEV image, which captures drivable areas, lane geometry, and vehicle locations and headings, and structured scene text describing vehicle states and ego-relative spatial and lane relations. The VLM-based scenario understanding agent is implemented using GPT-4o \cite{hurst2024gpt}.

The reasoning process comprises three stages. First, \emph{visual scene understanding} identifies drivable regions, lane layout and connectivity, and the lane-level locations of surrounding vehicles, thereby establishing the geometric constraints on feasible future motion. Second, \emph{agent behavior analysis} reasons about ego-relative interactions and plausible surrounding vehicle behaviors. Real-world crash-pattern statistics are incorporated as prior knowledge to guide behavior inference and candidate selection, while scene geometry and observed motion remain the primary evidence. Third, \emph{safety-critical interaction prediction} assesses whether the inferred future motions may create trajectory conflicts or insufficient safety margins with the ego vehicle. A feasibility check removes candidates that are unlikely to interact with the ego vehicle within the prediction horizon. The resulting candidate adversarial vehicles, potential adversarial behaviors, and structured scene-understanding description are used to construct the query for real-world crash-prior retrieval. Detailed VLM prompts, together with an example BEV image and its corresponding structured scene text, are provided in Supplementary Sections.

\subsection*{Real-world crash-prior retrieval and anchor construction}
\paragraph{Database construction.}
We constructed a structured real-world crash database from bird's-eye-view trajectory reconstructions of naturalistic crashes and near-crashes in the SHRP2 NDS~\cite{Jiao2026GSSM}. Each retained safety-critical interaction was assigned to one of 16 interaction modes derived from the NHTSA pre-crash scenario typology~\cite{najm2007precrash}. The trajectories were transformed into an ego-centred coordinate frame and resampled at 0.5\,s intervals. The processed interactions were organized into a three-layer database covering interaction semantics, individual crash events and temporal motion patterns. The mode layer represents each interaction class through its semantic description, vehicle manoeuvres and contextual factors. The event layer represents individual safety-critical events using event descriptions and event-level motion characteristics. The sample layer decomposes each event into 8\,s windows. The first 2\,s describe the observed motion, while the subsequent 6\,s target-vehicle trajectory provides the motion anchor for downstream generation. Moreover, natural-language descriptors at all three layers were encoded using the all-MiniLM-L6-v2 Sentence-Transformer model~\cite{reimers2019sentence,wang2020minilm}, providing a consistent semantic representation of the textual information associated with the real-world crash records.

The paired ego--target trajectories within the final 6\,s of each 8\,s window were automatically annotated with behavior-stage timings using a rule-based strategy. We first define a kinematic conflict condition as
\begin{equation}
R_j(\tau,\gamma,\alpha)
=
\left(
\mathrm{TTC}_j\leq\tau
\right)
\lor
\left(
c_j\geq\gamma
\right)
\lor
\left(
d_j\leq\alpha d_0
\right),
\end{equation}
where $d_j$, $c_j$ and $\mathrm{TTC}_j$ denote relative distance, closing speed and time to collision (TTC) at time step $j$, respectively. The onset indices of the risk-awareness, emergency-response and collision stages, denoted by $t_{\mathrm{risk}}$, $t_{\mathrm{emg}}$ and $t_{\mathrm{col}}$, respectively, are then defined as
\begin{equation}
\begin{aligned}
t_{\mathrm{risk}}
&=
\min
\left\{
j<t_{\mathrm{col}}:
R_j(\tau_r,\gamma_r,\alpha_r)
\right\},\\
t_{\mathrm{emg}}
&=
\min
\left\{
j<t_{\mathrm{col}}:
\left[
R_j(\tau_e,\gamma_e,\alpha_e)
\land
a_j\leq-\beta_e
\right]
\lor
a_j\leq-\beta_s
\right\},\\
t_{\mathrm{col}}
&=
\min
\left\{
j:
g_j\leq\varepsilon
\right\}.
\end{aligned}
\end{equation}
Here, $a_j$ is the target vehicle's longitudinal acceleration, $g_j$ is the minimum separation between the oriented bounding boxes, and $d_0$ is the initial relative distance. The positive parameters $\beta_e$ and $\beta_s$ denote the magnitudes of the emergency and strong-deceleration thresholds, respectively, with $\beta_s>\beta_e>0$. The parameters $\tau$, $\gamma$ and $\alpha$ specify the TTC, closing-speed and relative-distance thresholds. If no qualifying pre-collision time step existed, the corresponding stage onset was recorded as undefined. Details of the database construction are provided in Supplementary Section.

\paragraph{Prior retrieval and anchor construction.}
CrashSim implements the retrieval operator $\rho$ as a three-stage cascade over the crash database $\mathbf{d}_{\mathrm{crash}}$. Mode retrieval first identifies interaction classes that are semantically compatible with the query $\mathbf{q}$. Event retrieval then ranks individual events according to their interaction characteristics and motion consistency. Sample retrieval finally identifies the temporal motion pattern that best matches the observed motion.

Text embeddings are compared using cosine similarity, consistent with the Sentence-Transformer framework \cite{reimers2019sentence}. Cosine similarity is also applied to the signed event-feature and differential-motion vectors, with the resulting scores linearly rescaled to $[0,1]$ before fusion. Categorical, state and spatial similarities were normalized separately and combined using layer-specific weights.

Among the sample candidates retained after mode and event retrieval, CrashSim selects the candidate with the highest composite sample-level similarity score $S_{\mathrm{sample}}$. The matched crash prior $\mathbf{k}=\rho(\mathbf{q},\mathbf{d}_{\mathrm{crash}})$ is retained only when this score exceeds a predefined threshold $\delta$; otherwise, no crash-derived anchor is passed to the downstream trajectory generator.

The accepted crash prior $\mathbf{k}$ contains a 6 s differential-motion sequence,
\begin{equation}
\mathbf{k}_{\mathrm{motion}}
=
\left\{
\left(
\Delta \ell_j,
\Delta\theta_j
\right)
\right\}_{j=1}^{F},
\qquad
F=12,
\end{equation}
where $\Delta \ell_j$ and $\Delta\theta_j$ denote the distance increment and heading change over the $j$-th 0.5-s interval, respectively. Rather than directly transferring the absolute trajectory of the source crash, CrashSim preserves its relative-motion pattern and reconstructs the anchor trajectory from the current state $s^{i^{\star}}_{t}$ using midpoint integration:
\begin{equation}
\widetilde{\theta}_{j-\frac{1}{2}}
=
\widetilde{\theta}_{j-1}
+
\frac{1}{2}\Delta\theta_j,
\qquad
\widetilde{p}_j
=
\widetilde{p}_{j-1}
+
\Delta\ell_j
\begin{bmatrix}
\cos\widetilde{\theta}_{j-\frac{1}{2}}\\
\sin\widetilde{\theta}_{j-\frac{1}{2}}
\end{bmatrix},
\qquad
\widetilde{\theta}_j
=
\operatorname{wrap}
\left(
\widetilde{\theta}_{j-1}
+
\Delta\theta_j
\right),
\end{equation}
where
$\widetilde{p}_0
=(x^{i^{\star}}_t,y^{i^{\star}}_t)^{\mathsf T}$
and
$\widetilde{\theta}_0=\theta^{i^{\star}}_t$.
The reconstructed positions form the crash-informed anchor trajectory
$\mathbf{P}_{\mathrm{anc}}
=(\widetilde{p}_1,\ldots,\widetilde{p}_F)$.

The behavior-stage timings associated with the matched crash prior $\mathbf{k}$ are retrieved together with its motion pattern and collected as
$
\mathbf{t}_{\mathrm{stage}}
=
\left(
t_{\mathrm{risk}},
t_{\mathrm{emg}},
t_{\mathrm{col}}
\right).
$
The resulting anchor guidance is represented as
\begin{equation}
\mathbf{a}
=
\left(
i^{\star},
\mathbf{P}_{\mathrm{anc}},
\mathbf{t}_{\mathrm{stage}}
\right).
\end{equation}
This construction preserves the temporal motion pattern and behavior-stage progression of the matched real-world crash while spatially aligning the reconstructed anchor trajectory with the current state of the candidate adversarial vehicle. Further implementation details are provided in Supplementary Section.

\subsection*{Diffusion-based crash scenario generation}

CrashSim uses a scene-conditioned latent diffusion model to generate the future trajectories $\boldsymbol{\tau}^{1:N-1}_{t:t+T}$ of the surrounding vehicles. Given the road map and observed multi-agent motion history $(\mathbf{m},\mathbf{x})$, the encoder $E$ produces a scene-conditioning latent $\mathbf{c}=E(\mathbf{m},\mathbf{x})$. Both the encoder and decoder adopt interaction-aware graph architectures to model dependencies among traffic participants~\cite{Rempe2022,Peng2026LDScene}. The denoising network adopts a temporal U-Net composed of residual temporal blocks and linear-attention layers~\cite{janner2022planning}. The diffusion model is trained on the nuScenes training split using the standard noise-prediction objective~\cite{ho2020denoising}, with the denoising network conditioned on the scene context $\mathbf{c}$. Model architecture and training details are provided in Supplementary Section.

At inference, sampling is initialized from Gaussian noise, $\mathbf{z}^{K}\sim\mathcal{N}(\mathbf{0},\mathbf{I})$, and progressively denoises the latent through the DDIM reverse process~\cite{song2020denoising}:
\begin{equation}
\mathbf{z}^{k-1}
=
\sqrt{\bar{\alpha}_{k-1}}
\left(
\frac{
\mathbf{z}^{k}
-
\sqrt{1-\bar{\alpha}_{k}}\,
\boldsymbol{\epsilon}_{\theta}
\left(
\mathbf{z}^{k},k,\mathbf{c}
\right)
}{
\sqrt{\bar{\alpha}_{k}}
}
\right)
+
\sqrt{1-\bar{\alpha}_{k-1}}\,
\boldsymbol{\epsilon}_{\theta}
\left(
\mathbf{z}^{k},k,\mathbf{c}
\right),
\label{denoising_process}
\end{equation}
where $\bar{\alpha}_{k}$ denotes the cumulative noise schedule and $\boldsymbol{\epsilon}_{\theta}$ denotes the scene-conditioned denoising network that predicts the noise component at diffusion step $k$. The reverse process iteratively removes noise from $\mathbf{z}^{K}$ to obtain the clean latent $\mathbf{z}^{0}$. Finally, the decoder $D$ maps $\mathbf{z}^{0}$, together with the scene context, to the joint future trajectories $\boldsymbol{\tau}^{1:N-1}_{t:t+T}$ of the surrounding vehicles. The decoder further incorporates autoregressive motion prediction and kinematic-bicycle-model propagation to produce physically continuous and dynamically feasible multi-agent trajectories \cite{Rempe2022}.

A key component of CrashSim is \emph{human behavior-informed anchor guidance}, which integrates the temporal interaction structure of real-world crashes into the generative process to facilitate the generation of realistic crash trajectories. The anchor $\mathbf{a}$ comprises spatial waypoints and behavior-stage annotations, which jointly guide the diffusion process. This representation preserves characteristic human responses preceding and during a crash while allowing the generated trajectory to adapt to the geometry and traffic configuration of the current scene.

During the final few reverse diffusion steps, we apply guidance based on the clean-guidance strategy~\cite{rempe2023trace}. At each guided step, the predicted clean latent $\hat{\mathbf{z}}^{0}$ is decoded into trajectory space, where a differentiable guidance loss $J$ is computed based on the decoded trajectories and the anchor guidance $\mathbf{a}$. Following gradient-based diffusion guidance~\cite{dhariwal2021diffusion}, the noise prediction is modified as
\begin{equation}
\widetilde{\boldsymbol{\epsilon}}_{\theta}
\left(
\mathbf{z}^{k},k,\mathbf{c}
\right)
=
\boldsymbol{\epsilon}_{\theta}
\left(
\mathbf{z}^{k},k,\mathbf{c}
\right)
-
s\,
\nabla_{\mathbf{z}^{k}}
J
\left[
D_{\theta}
\left(
\hat{\mathbf{z}}^{0},
\mathbf{x},
\mathbf{m}
\right);
\mathbf{a}
\right],
\label{eq:guided_noise}
\end{equation}
where $s$ denotes the guidance scale. During the guided steps, $\widetilde{\boldsymbol{\epsilon}}_{\theta}$ replaces $\boldsymbol{\epsilon}_{\theta}$ in the DDIM update in Eq.~(\ref{denoising_process}) to obtain $\mathbf{z}^{k-1}$.

The guidance loss combines behavioral anchoring, stage consistency, safety-critical interaction, trajectory realism and route consistency:
\begin{equation}
J
=
w_{\mathrm{anchor}}J_{\mathrm{anchor}}
+
w_{\mathrm{stage}}J_{\mathrm{stage}}
+
w_{\mathrm{coll}}J_{\mathrm{coll}}
+
w_{\mathrm{real}}J_{\mathrm{real}}
+
w_{\mathrm{route}}J_{\mathrm{route}},
\label{eq:guidance_loss}
\end{equation}
where the weights control the relative contributions of the five guidance terms.

The anchor guidance term transfers the spatial progression and response dynamics of the retrieved crash anchor to the candidate adversarial vehicle \(i^{\star}\):
\begin{equation}
\begin{aligned}
J_{\mathrm{anchor}}
={}&
\left\langle
w_{p}
\rho\!\left(
\left\|
p_{t}-\bar{p}_{t}
\right\|_{2}
\right)
+
w_{a}
\left(a_{t}-\bar{a}_{t}\right)^{2}
+
w_{\psi}
\operatorname{wrap}\!\left(
\Delta\psi_{t}-\Delta\bar{\psi}_{t}
\right)^{2}
\right\rangle_{t\in\mathcal{V}},
\end{aligned}
\label{eq:anchor_loss}
\end{equation}
where \(p_{t}\), \(a_{t}\) and \(\Delta\psi_{t}\) denote position, longitudinal acceleration and heading change, respectively. Barred quantities are computed from the anchor, and \(\rho\) denotes the smooth-\(L_{1}\) penalty. The set \(\mathcal{V}\) contains pre-impact anchor waypoints and excludes off-road segments occurring after the anchor has entered the drivable area. This treatment preserves supervision at initial spawn locations while preventing incompatible post-entry segments from being reproduced.

The stage guidance term aligns the generated interaction with the risk-awareness,
emergency-response and collision stages encoded by the retrieved anchor.
Using the stage timings defined above, let
$t_{\mathrm{risk}}$, $t_{\mathrm{emg}}$ and
$t_{\mathrm{col}}$ denote the corresponding timings inferred
from the generated interaction, while
$t_{\mathrm{risk}}^\star$, $t_{\mathrm{emg}}^\star$ and
$t_{\mathrm{col}}^\star$ denote those of the anchor. The stage objective is

\begin{equation}
\begin{aligned}
J_{\mathrm{stage}}
={}&
w_r
\left(
t_{\mathrm{risk}}-t_{\mathrm{risk}}^\star
\right)^{2}
+ w_m
\left(
t_{\mathrm{emg}}-t_{\mathrm{emg}}^\star
\right)^{2}
+ w_c
\left(
t_{\mathrm{col}}-t_{\mathrm{col}}^\star
\right)^{2}
+w_d E_{\mathrm{dur}}
+w_o E_{\mathrm{ord}},
\end{aligned}
\label{eq:stage_loss}
\end{equation}

The first three terms align the onset of risk awareness, emergency response
and collision with the retrieved crash anchor, while $E_{\mathrm{dur}}$
discourages shortening of the emergency-response-to-collision interval and
$E_{\mathrm{ord}}$ penalizes violations of the expected
risk-awareness--emergency-response-collision ordering. In implementation,
the timing and duration errors are normalized by the prediction horizon to
ensure a consistent scale across trajectories, and only terms with valid stage
annotations are included. The stage timings are obtained using differentiable
soft versions of the behavior-stage criteria defined above, with the detailed
normalization and weighting coefficients provided
in the Supplementary Section.

Together, $J_{\mathrm{anchor}}$ and $J_{\mathrm{stage}}$ transfer human response patterns from real-world crash processes into the generated interaction.

The collision term $J_{\mathrm{coll}}$ promotes a safety-critical interaction between adversarial vehicle $i^{\star}$ and the ego vehicle, while the route term $J_{\mathrm{route}}$ encourages $i^{\star}$ to follow a plausible route consistent with the reference motion pattern. The realism term $J_{\mathrm{real}}$ penalizes off-road motion and unintended collisions. Following LD-scene~\cite{Peng2026LDScene}, role-specific gradient masks apply all five guidance terms to the adversarial vehicle $i^{\star}$, while only the realism term $J_{\mathrm{real}}$ is applied to the remaining surrounding vehicles; no gradient is applied to the ego trajectory. Overall, this guidance formulation preserves variation in interaction timing, conflict geometry and vehicle responses while retaining the behavioral characteristics of real-world crash processes.

For each scene-anchor pair, CrashSim samples multiple latent realizations and retains the candidate with the lowest guidance score after penalizing violations of physical feasibility. Details of guidance terms are provided in Supplementary Section.

\subsection*{LLM-assisted planner evaluation}

We develop an LLM-assisted evaluation framework to characterize how planner capabilities change from naturalistic driving in nuScenes to safety-critical interactions in nuCrash (Fig.~\ref{fig:method}b). nuCrash is generated by CrashSim to introduce human behavior-informed safety-critical interactions while preserving the scene and trajectory representation used for nuScenes. Each planner is evaluated on both datasets under the same closed-loop rollout protocol, with its behavior quantified using deterministic metrics and scenario-level statistics covering safety, response timing, longitudinal and lateral control, comfort and stability, and mobility efficiency.

While these measurements quantify individual aspects of planner performance, they remain fragmented across metrics, scenarios and failure patterns and do not directly reveal the underlying capability limitations. We therefore introduce an LLM-assisted evaluation agent, implemented using GPT-4o~\cite{hurst2024gpt}, to analyze this heterogeneous evidence through evidence analysis and structured reasoning~\cite{shi2025evidence,qiu2025quantifying}. For each dataset, the agent jointly interprets metric outcomes, failure patterns and their scenario contexts to derive a capability profile along six dimensions: risk anticipation, interaction reasoning, longitudinal response, lateral response, comfort and stability, and mobility efficiency. This analysis identifies consistent behavioral limitations and the interaction conditions under which they emerge, rather than relying on any individual metric as a direct indicator of planner capability.

The evaluation agent then performs cross-dataset analysis by contrasting the capability profiles obtained from nuScenes and nuCrash. This comparison distinguishes limitations already present in naturalistic driving from those amplified or newly exposed under safety-critical interactions. By relating capability changes to the corresponding metric shifts, failure patterns and scenario contexts, the agent identifies the associated failure mechanisms, attributes them to underlying behavioral limitations, and translates the resulting diagnosis into targeted directions for planner improvement and subsequent validation. Detailed metric definitions and prompts are provided in Supplementary Section.

\newpage
\bibliography{ref}

@string{CVPR = {the Proceedings of the IEEE/CVF Conference on Computer Vision and Pattern Recognition (CVPR)}}

@string{ECCV = {the Proceedings of European Conference on Computer Vision (ECCV)}}

@string{CORL = {the Proceedings of Conference on Robot Learning (CoRL)}}

@string{ICLR = {the Proceedings of International Conference on Learning Representations (ICLR)}}

@string{NIPS = {the Proceedings of Advances in Neural Information Processing Systems}}

@string{ICML = {the Proceedings of International Conference on Machine Learning (ICML)}}

@string{ITSC = {IEEE International Conference on Intelligent Transportation Systems (ITSC)}}

@article{Feng2021,
  author  = {Feng, Shuo and Yan, Xintao and Sun, Haowei and Feng, Yiheng and Liu, Henry X.},
  title   = {Intelligent driving intelligence test for autonomous vehicles with naturalistic and adversarial environment},
  journal = {Nature Communications},
  volume  = {12},
  pages   = {748},
  year    = {2021},
}

@article{Yan2023,
  author  = {Yan, Xintao and Zou, Zhengxia and Feng, Shuo and Zhu, Haojie and Sun, Haowei and Liu, Henry X.},
  title   = {Learning naturalistic driving environment with statistical realism},
  journal = {Nature Communications},
  volume  = {14},
  pages   = {2037},
  year    = {2023},
}

@inproceedings{Rempe2022,
  author    = {Rempe, Davis and Philion, Jonah and Guibas, Leonidas J. and Fidler, Sanja and Litany, Or},
  title     = {Generating useful accident-prone driving scenarios via a learned traffic prior},
  booktitle = CVPR,
  pages     = {17284--17294},
  year      = {2022},
  organization={IEEE}
}

@article{Peng2026LDScene,
  author  = {Peng, Mingxing and Xie, Yuting and Guo, Xusen and Yao, Ruoyu and Yang, Hai and Ma, Jun},
  title   = {{LD-scene}: {LLM}-guided diffusion for controllable generation of adversarial safety-critical driving scenarios},
  journal = {Transportation Research Part C: Emerging Technologies},
  volume  = {188},
  pages   = {105694},
  year    = {2026}
}

@inproceedings{Peng2025GuidedLDM,
  author    = {Peng, Mingxing and Yao, Ruoyu and Guo, Xusen and Xie, Yuting and Chen, Xianda and Ma, Jun},
  title     = {Safety-critical traffic simulation with guided latent diffusion model},
  booktitle = ITSC,
  pages     = {987--993},
  year      = {2025},
  organization = {IEEE}
}

@article{antin2019second,
  title={Second strategic highway research program naturalistic driving study methods},
  author={Antin, Jonathan F and Lee, Suzie and Perez, Miguel A and Dingus, Thomas A and Hankey, Jonathan M and Brach, Ann},
  journal={Safety Science},
  volume={119},
  pages={2--10},
  year={2019},
  publisher={Elsevier}
}

@techreport{Hankey2016SHRP2,
  author      = {Hankey, Jonathan M. and Perez, Miguel A. and McClafferty, Julie A.},
  title       = {Description of the SHRP 2 naturalistic database and the crash, near-crash, and baseline data sets},
  institution = {Virginia Tech Transportation Institute},
  year        = {2016},
  number      = {SHRP2-S2-S06-RR-1},
  url         = {https://vtechworks.lib.vt.edu/bitstream/handle/10919/70850/SHRP_2_CrashNearCrashBaselineReport_4-25-16.pdf}
}

@techreport{najm2007precrash,
  title     = {Pre-Crash Scenario Typology for Crash Avoidance Research},
  author    = {Najm, Wassim G. and Smith, John D. and Yanagisawa, Mikio},
  institution = {National Highway Traffic Safety Administration},
  number    = {DOT HS 810 767},
  year      = {2007},
}

@article{Jiao2026GSSM,
  author  = {Jiao, Yiru and Calvert, Simeon C. and van Cranenburgh, Sander and van Lint, Hans},
  title   = {Learning collision risk proactively from naturalistic driving data at scale},
  journal = {Nature Machine Intelligence},
  volume  = {8},
  number  = {3},
  pages   = {337--350},
  year    = {2026},
}

@inproceedings{Lewis2020RAG,
  author    = {Lewis, Patrick and Perez, Ethan and Piktus, Aleksandra and Petroni, Fabio and Karpukhin, Vladimir and Goyal, Naman and Kuttler, Heinrich and Lewis, Mike and Yih, Wen-tau and Rocktaschel, Tim and Riedel, Sebastian and Kiela, Douwe},
  title     = {Retrieval-augmented generation for knowledge-intensive {NLP} tasks},
  booktitle = NIPS,
  volume    = {33},
  pages     = {9459--9474},
  year      = {2020}
}

@inproceedings{nuscenes2019,
  title={{nuScenes}: A multimodal dataset for autonomous driving},
  author={Caesar, Holger and Bankiti, Varun and Lang, Alex H and Vora, Sourabh and Liong, Venice Erin and Xu, Qiang and others},
  booktitle=CVPR,
  pages={11621--11631},
  year={2020},
  organization={IEEE}
}

@article{li2026world,
  title={{World Engine}: Towards the Era of Post-Training for Autonomous Driving},
  author={Li, Tianyu and Chen, Li and Wang, Caojun and Liu, Haochen and Chitta, Kashyap and Yang, Zhenjie and Lu, Yuhang and Ye, Naisheng and Qiu, Yihang and Wang, Yufei and others},
  journal={arXiv preprint arXiv:2606.19836},
  year={2026},
  doi={https://arxiv.org/abs/2606.19836}
}

@article{gibbs2002choosing,
  title={On choosing and bounding probability metrics},
  author={Gibbs, Alison L and Su, Francis Edward},
  journal={International Statistical Review},
  volume={70},
  number={3},
  pages={419--435},
  year={2002},
  publisher={Wiley Online Library}
}

@article{lin1991divergence,
  title={Divergence measures based on the Shannon entropy},
  author={Lin, Jianhua},
  journal={IEEE Transactions on Information Theory},
  volume={37},
  number={1},
  pages={145--151},
  year={1991},
  publisher={IEEE}
}

@article{montemerlo2008junior,
  title={{Junior}: The stanford entry in the urban challenge},
  author={Montemerlo, Michael and Becker, Jan and Bhat, Suhrid and Dahlkamp, Hendrik and Dolgov, Dmitri and Ettinger, Scott and Haehnel, Dirk and Hilden, Tim and Hoffmann, Gabe and Huhnke, Burkhard and others},
  journal={Journal of Field Robotics},
  volume={25},
  number={9},
  pages={569--597},
  year={2008},
}

@inproceedings{dauner2023parting,
  title={Parting with misconceptions about learning-based vehicle motion planning},
  author={Dauner, Daniel and Hallgarten, Marcel and Geiger, Andreas and Chitta, Kashyap},
  booktitle=CoRL,
  pages={1268--1281},
  year={2023},
}

@article{bansal2018chauffeurnet,
  title={{ChauffeurNet}: Learning to drive by imitating the best and synthesizing the worst},
  author={Bansal, Mayank and Krizhevsky, Alex and Ogale, Abhijit},
  journal={arXiv preprint arXiv:1812.03079},
  doi={https://arxiv.org/abs/1812.03079},
  year={2018}
}

@inproceedings{suo2021trafficsim,
  title={{TrafficSim}: Learning to simulate realistic multi-agent behaviors},
  author={Suo, Simon and Regalado, Sebastian and Casas, Sergio and Urtasun, Raquel},
  booktitle=CVPR,
  pages={10400--10409},
  year={2021},
  organization={IEEE}
}

@inproceedings{ICLR2025_5c0cd430,
 author = {Zheng, Yinan and Liang, Ruiming and ZHENG, Kexin and Zheng, Jinliang and Mao, Liyuan and Li, Jianxiong and Gu, Weihao and Ai, Rui and Li, Shengbo and Zhan, Xianyuan and Liu, Jingjing},
 booktitle = ICLR,
 pages = {37207--37227},
 title = {Diffusion-Based Planning for Autonomous Driving with Flexible Guidance},
 volume = {2025},
 year = {2025}
}

@article{njord2015bigdata,
  title   = {Big Data Hit the Road: The First Year of Use of the {SHRP 2} Safety Databases},
  author  = {Njord, John and Steudle, Kirk},
  journal = {TR News},
  number  = {300},
  pages   = {3--8},
  year    = {2015},
  month   = {November--December},
  publisher = {Transportation Research Board}
}

@inproceedings{reimers2019sentence,
  title={{Sentence-BERT}: Sentence embeddings using siamese bert-networks},
  author={Reimers, Nils and Gurevych, Iryna},
  booktitle={the Proceedings of the 2019 Conference on Empirical Methods in Natural Language Processing and the 9th International Joint Conference on Natural Language Processing},
  pages={3982--3992},
  year={2019}
}

@article{wang2020minilm,
  title={{MiniLM}: Deep self-attention distillation for task-agnostic compression of pre-trained transformers},
  author={Wang, Wenhui and Wei, Furu and Dong, Li and Bao, Hangbo and Yang, Nan and Zhou, Ming},
  journal=NIPS,
  volume={33},
  pages={5776--5788},
  year={2020}
}

@article{hurst2024gpt,
  title={{Gpt-4o} system card},
  author={Hurst, Aaron and Lerer, Adam and Goucher, Adam P and Perelman, Adam and Ramesh, Aditya and Clark, Aidan and Ostrow, AJ and Welihinda, Akila and Hayes, Alan and Radford, Alec and others},
  journal={arXiv preprint arXiv:2410.21276},
  doi={https://arxiv.org/abs/2410.21276},
  year={2024}
}

@article{song2020denoising,
  title={Denoising diffusion implicit models},
  author={Song, Jiaming and Meng, Chenlin and Ermon, Stefano},
  journal={arXiv preprint arXiv:2010.02502},
  year={2020},
  doi={https://arxiv.org/abs/2010.02502}
}

@inproceedings{janner2022planning,
  title={Planning with diffusion for flexible behavior synthesis},
  author={Janner, Michael and Du, Yilun and Tenenbaum, Joshua B and Levine, Sergey},
  booktitle=ICML,
  pages={9902--9915},
  year={2022}
}

@inproceedings{rempe2023trace,
  title={{Trace and Pace}: Controllable pedestrian animation via guided trajectory diffusion},
  author={Rempe, Davis and Luo, Zhengyi and Peng, Xuebin and Yuan, Ye and Kitani, Kris and Kreis, Karsten and Fidler, Sanja and Litany, Or},
  booktitle=CVPR,
  pages={13756--13766},
  year={2023},
  organization = {IEEE}
}

@article{dhariwal2021diffusion,
  title={Diffusion models beat gans on image synthesis},
  author={Dhariwal, Prafulla and Nichol, Alexander},
  journal=NIPS,
  volume={34},
  pages={8780--8794},
  year={2021}
}

@article{shi2025evidence,
  title={Evidence triangulator: using large language models to extract and synthesize causal evidence across study designs},
  author={Shi, Xuanyu and Zhao, Wenjing and Chen, Ting and Yang, Chao and Du, Jian},
  journal={Nature Communications},
  volume={16},
  number={1},
  pages={7355},
  year={2025},
  publisher={Nature Publishing Group UK London}
}

@article{qiu2025quantifying,
  title={Quantifying the reasoning abilities of LLMs on clinical cases},
  author={Qiu, Pengcheng and Wu, Chaoyi and Liu, Shuyu and Fan, Yanjie and Zhao, Weike and Chen, Zhuoxia and Gu, Hongfei and Peng, Chuanjin and Zhang, Ya and Wang, Yanfeng and others},
  journal={Nature Communications},
  volume={16},
  number={1},
  pages={9799},
  year={2025},
  publisher={Nature Publishing Group UK London}
}

@article{ho2020denoising,
  title={Denoising diffusion probabilistic models},
  author={Ho, Jonathan and Jain, Ajay and Abbeel, Pieter},
  journal=NIPS,
  volume={33},
  pages={6840--6851},
  year={2020}
}

@article{schumann2026active,
  title={Active inference as a model of collision avoidance behavior in human drivers},
  author={Schumann, Julian F and Engstr{\"o}m, Johan and Johnson, Leif and O’Kelly, Matthew and Messias, Joao and Kober, Jens and Zgonnikov, Arkady},
  journal={Nature Communications},
  volume={17},
  number={1},
  pages={5009},
  year={2026},
  publisher={Nature Publishing Group}
}

@article{qian2026test,
  title={Test case sampling optimization for safety validation of automated driving systems},
  author={Qian, Chen and Xu, Jingbin and Xing, Xin and Guo, Feng},
  journal={Nature Communications},
  volume={17},
  number={1},
  pages={3114},
  year={2026},
  publisher={Nature Publishing Group UK London}
}

@article{feng2023dense,
  title={Dense reinforcement learning for safety validation of autonomous vehicles},
  author={Feng, Shuo and Sun, Haowei and Yan, Xintao and Zhu, Haojie and Zou, Zhengxia and Shen, Shengyin and Liu, Henry X},
  journal={Nature},
  volume={615},
  number={7953},
  pages={620--627},
  year={2023},
  publisher={Nature Publishing Group UK London}
}

@inproceedings{wang2021advsim,
  title={{AdvSim}: Generating safety-critical scenarios for self-driving vehicles},
  author={Wang, Jingkang and Pun, Ava and Tu, James and Manivasagam, Sivabalan and Sadat, Abbas and Casas, Sergio and Ren, Mengye and Urtasun, Raquel},
  booktitle=CVPR,
  pages={9909-9918},
  year={2021},
  organization = {IEEE}
}

@inproceedings{hanselmann2022king,
  title={{KING}: Generating safety-critical driving scenarios for robust imitation via kinematics gradients},
  author={Hanselmann, Niklas and Renz, Katrin and Chitta, Kashyap and Bhattacharyya, Apratim and Geiger, Andreas},
  booktitle=ECCV,
  pages={335--352},
  year={2022},
  organization={Springer}
}

@article{svard2021computational,
  title={Computational modeling of driver pre-crash brake response, with and without off-road glances: Parameterization using real-world crashes and near-crashes},
  author={Sv{\"a}rd, Malin and Markkula, Gustav and B{\"a}rgman, Jonas and Victor, Trent},
  journal={Accident Analysis \& Prevention},
  volume={163},
  pages={106433},
  year={2021},
  publisher={Elsevier}
}

@article{guo2023modeling,
  title={Modeling driver’s evasive behavior during safety--critical lane changes: Two-dimensional time-to-collision and deep reinforcement learning},
  author={Guo, Hongyu and Xie, Kun and Keyvan-Ekbatani, Mehdi},
  journal={Accident Analysis \& Prevention},
  volume={186},
  pages={107063},
  year={2023},
  publisher={Elsevier}
}

@article{Kalra2016Driving,
  title={Driving to safety: How many miles of driving would it take to demonstrate autonomous vehicle reliability?},
  author={Kalra, Nidhi and Paddock, Susan M},
  journal={Transportation Research Part A: Policy and Practice},
  volume={94},
  pages={182--193},
  year={2016},
  publisher={Elsevier}
}

@article{liu2024curse,
  title={Curse of rarity for autonomous vehicles},
  author={Liu, Henry X and Feng, Shuo},
  journal={Nature Communications},
  volume={15},
  number={1},
  pages={4808},
  year={2024},
  publisher={Nature Publishing Group UK London}
}

@article{tafidis2022safety,
  title={Safety implications of higher levels of automated vehicles: a scoping review},
  author={Tafidis, Pavlos and Farah, Haneen and Brijs, Tom and Pirdavani, Ali},
  journal={Transport reviews},
  volume={42},
  number={2},
  pages={245--267},
  year={2022}
}

@article{sohrabi2021quantifying,
  title={Quantifying the automated vehicle safety performance: A scoping review of the literature, evaluation of methods, and directions for future research},
  author={Sohrabi, Soheil and Khodadadi, Ali and Mousavi, Seyedeh Maryam and Dadashova, Bahar and Lord, Dominique},
  journal={Accident Analysis \& Prevention},
  volume={152},
  pages={106003},
  year={2021},
  publisher={Elsevier}
}

@article{ding2023survey,
  author={Ding, Wenhao and Xu, Chejian and Arief, Mansur and Lin, Haohong and Li, Bo and Zhao, Ding},
  journal={IEEE Transactions on Intelligent Transportation Systems}, 
  title={A Survey on Safety-Critical Driving Scenario Generation—A Methodological Perspective}, 
  year={2023},
  volume={24},
  number={7},
  pages={6971-6988},
}

@inproceedings{dosovitskiy2017carla,
  title={{CARLA}: An open urban driving simulator},
  author={Dosovitskiy, Alexey and Ros, German and Codevilla, Felipe and Lopez, Antonio and Koltun, Vladlen},
  booktitle=CoRL,
  pages={1--16},
  year={2017},
}

@inproceedings{lopez2018microscopic,
  title={Microscopic traffic simulation using sumo},
  author={Lopez, Pablo Alvarez and Behrisch, Michael and Bieker-Walz, Laura and Erdmann, Jakob and Fl{\"o}tter{\"o}d, Yun-Pang and Hilbrich, Robert and L{\"u}cken, Leonhard and Rummel, Johannes and Wagner, Peter and Wie{\ss}ner, Evamarie},
  booktitle=ITSC,
  pages={2575--2582},
  year={2018},
}

@inproceedings{chang2024safe,
  title={{SAFE-SIM}: Safety-Critical Closed-Loop Traffic Simulation with Diffusion-Controllable Adversaries},
  author={Chang, Wei-Jer and Pittaluga, Francesco and Tomizuka, Masayoshi and Zhan, Wei and Chandraker, Manmohan},
  booktitle=ECCV,
  pages={242--258},
  year={2024},
  organization={Springer}
}

@article{wu2026vlm,
  title={{VLM as strategist}: Adaptive generation of safety-critical testing scenarios via guided diffusion},
  author={Wu, Xinzheng and Chen, Junyi and Zhong, Naiting and Shen, Yong},
  journal={Accident Analysis \& Prevention},
  volume={236},
  pages={108680},
  year={2026},
  publisher={Elsevier}
}

@article{mei2025llm,
  title={{LLM-Attacker}: Enhancing closed-loop adversarial scenario generation for autonomous driving with large language models},
  author={Mei, Yuewen and Nie, Tong and Sun, Jian and Tian, Ye},
  journal={IEEE Transactions on Intelligent Transportation Systems},
  volume={26},
  number={10},
  pages={15068--15076},
  year={2025},
  publisher={IEEE}
}

@article{chang2024llmscenario,
  title={{LLMscenario}: Large language model driven scenario generation},
  author={Chang, Cheng and Wang, Siqi and Zhang, Jiawei and Ge, Jingwei and Li, Li},
  journal={IEEE Transactions on Systems, Man, and Cybernetics: Systems},
  volume={54},
  number={11},
  pages={6581--6594},
  year={2024},
}

@article{jiang2026anchordrive,
  title={{AnchorDrive}: LLM Scenario Rollout with Anchor-Guided Diffusion Regeneration for Safety-Critical Scenario Generation},
  author={Jiang, Zhulin and Li, Zetao and Wang, Cheng and Wang, Ziwen and Xiong, Chen},
  journal={arXiv preprint arXiv:2603.02542},
  year={2026},
  doi={https://arxiv.org/abs/2603.02542}
}

@article{nunes2018people,
  title={People must retain control of autonomous vehicles},
  author={Nunes, Ashley and Reimer, Bryan and Coughlin, Joseph F},
  journal={Nature},
  volume={556},
  number={7700},
  pages={169--171},
  year={2018},
  publisher={Nature Publishing Group UK London}
}

@article{yao2022dota,
  title={{DoTA}: Unsupervised detection of traffic anomaly in driving videos},
  author={Yao, Yu and Wang, Xizi and Xu, Mingze and Pu, Zelin and Wang, Yuchen and Atkins, Ella and Crandall, David J},
  journal={IEEE Transactions on Pattern Analysis and Machine Intelligence},
  volume={45},
  number={1},
  pages={444--459},
  year={2022},
  publisher={IEEE}
}

@article{fang2021dada,
  title={{DADA}: Driver attention prediction in driving accident scenarios},
  author={Fang, Jianwu and Yan, Dingxin and Qiao, Jiahuan and Xue, Jianru and Yu, Hongkai},
  journal={IEEE Transactions on Intelligent Transportation Systems},
  volume={23},
  number={6},
  pages={4959--4971},
  year={2021},
  publisher={IEEE}
}

@article{caesar2021nuplan,
  title={{NuPlan}: A closed-loop ml-based planning benchmark for autonomous vehicles},
  author={Caesar, Holger and Kabzan, Juraj and Tan, Kok Seang and Fong, Whye Kit and Wolff, Eric and Lang, Alex and Fletcher, Luke and Beijbom, Oscar and Omari, Sammy},
  journal={arXiv preprint arXiv:2106.11810},
  year={2021},
  doi={https://arxiv.org/abs/2106.11810}
}

@inproceedings{peng2026nuplan,
  title={{nuPlan-R}: A closed-loop planning benchmark for autonomous driving via reactive multi-agent simulation},
  author={Peng, Mingxing and Yao, Ruoyu and Guo, Xusen and Ma, Jun},
  booktitle={IEEE Intelligent Vehicles Symposium},
  pages={702--709},
  year={2026},
  organization={IEEE}
}

\label{LastMainPage}

\end{document}